\documentclass{TEAI}
\pdfoutput=1

\usepackage{enumitem}
\usepackage{flafter}     

\usepackage{algorithm}     
\usepackage{algorithmic}   
\usepackage{array}
\usepackage{tikz}
\usetikzlibrary{arrows.meta,positioning,fit,backgrounds}
\graphicspath{{figures/}}

\newcommand{\jspace}{J-space}
\newcommand{\jsf}{J64}
\newcommand{\pp}{\text{pp}}
\newcommand{\sig}{$^{*}$}

\definecolor{gain}{rgb}{0.10,0.52,0.16}
\definecolor{loss}{rgb}{0.72,0.11,0.11}
\newcommand{\gp}[1]{\,\textcolor{gain}{\scriptsize(+#1)}}
\newcommand{\gm}[1]{\,\textcolor{loss}{\scriptsize($-$#1)}}
\newcommand{\gz}[1]{\,\textcolor{loss!45!black}{\scriptsize(#1)}}

\title{Beyond the Trace: Coupling an Interpretable Reasoning-State Readout to Native MoE Routing}

\author[1*]{Kang Chen}
\author[1*]{Sihan Zhao}
\author[1,2\dagger]{Yixin Cao}
\author[1]{Yu-Gang Jiang}
\affiliation[1]{Fudan University}
\affiliation[2]{Shanghai Innovation Institute}
\checkdata[Email]{\email{kchen24@m.fudan.edu.cn}, \email{yxcao@fudan.edu.cn}$^{\dagger}$}
\newcommand{\jarurl}{https://cckfdu.com/jar/}
\checkdata[Project]{\href{\jarurl}{\texttt{cckfdu.com/jar}}}

\abstract{
What a reasoning model writes is only a partial record of the process that produces
it. We introduce a two-level internal readout for mixture-of-experts reasoning. We
first distill vocabulary-scale \jspace{} into \jsf{}, a $64$-axis semantic frame
learned from the model's own reasoning states. \jsf{} reveals readable process state
that the emitted trace does not show: it separates inference effort from
problem-induced strain. It also adds $0.096$ to $0.135$ held-out AUC over a
baseline that reads the same rollout as token occupancy and aggregates it in exactly
the same way. We then reconstruct \jsf{} from native expert-routing statistics. The result is
R64, a low-overhead proxy: its median per-axis correlation with \jsf{} is $0.69$ to
$0.86$ across three models and two families, and on gpt-oss-20b it preserves $95$ to
$100\%$ of \jsf{}'s predictive gain. The readout supports test-time decisions at two temporal
resolutions. Over completed candidate sets, \jsf{} and R64 improve single-branch
selection, and R64-weighted voting improves plain majority voting in seven of eight
settings. During generation, rolling
readout windows drive a cumulative stop-and-resample policy whose operating point is
fixed on training questions alone. \jsf{} improves accuracy by $1.1$ to $5.9$ points
over a sibling-permuted control, and the routing-only R64 proxy retains $0.9$ to $3.2$ of those points.
Finally, router edits aimed at the mechanism \jsf{} names induce the predicted reasoning behaviors and
shift a diagnosed stall from numerical guessing toward exact symbolic execution.
Together, \jsf{} makes latent process state readable, while routing makes it
deployable and actionable.
}

\begin{document}
\maketitle

\section{Introduction}

\begin{figure*}[t!]
\centering
\includegraphics{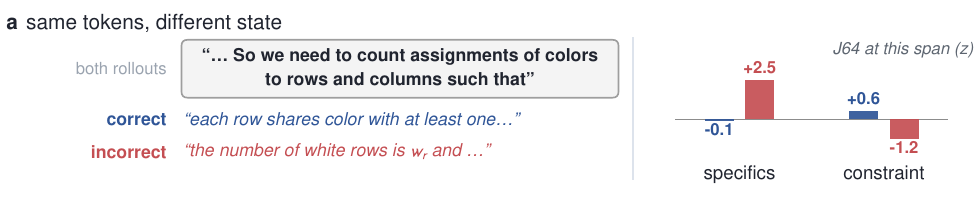}\\[4pt]
\includegraphics{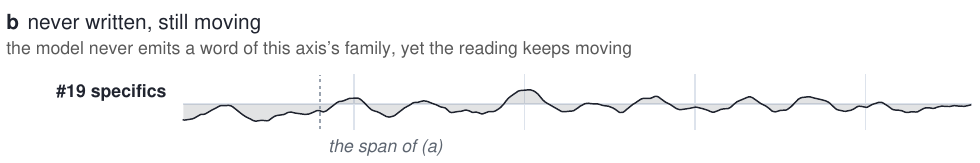}\\[4pt]
\includegraphics{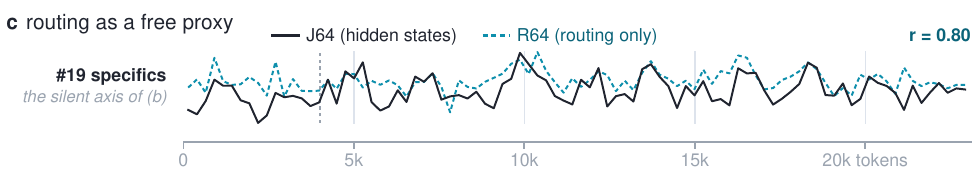}
\caption{\textbf{The internal readout at a glance, on one gpt-oss-20b High rollout of
AIME-24 problem~$29$.}
\textbf{(a)}~A pair matched on the emitted text: rollout~$1898$ and a sibling rollout on the same
problem emit an identical $15$-token span, yet their \jsf{} readings over that
span separate along named axes (bars, standardized within the problem). The
continuations diverge accordingly, and only one rollout is correct.
\textbf{(b)}~The same rollout end to end on the \emph{specifics} axis of (a)
($1024$-token centered average). An axis's \emph{token family} is its nearest
$20$ vocabulary directions. Not one of this axis's family words is written in the
$23$k tokens, and the same holds for $31$ of the $64$ axes, yet the reading keeps
moving.
\textbf{(c)}~The silent axis of (b), reconstructed from native routing alone and plotted on
the same token axis. A ridge map, fitted with question-held-out folds, recovers it from
the layer-$20$ expert usage spectrum of $256$-token windows. Median per-axis $r$ is
$0.69$ over the $64$ axes with $30$ above $0.7$ ($0.00$ for shuffled routing). The axis
shown ranks $12$th.}
\label{fig:teaser}
\end{figure*}

Reasoning traces are the primary interface for inspecting and controlling
reasoning models, but they record what a model chooses to emit rather than the
full process state that produces it. They do not directly reveal which
constraints and alternatives remain active, whether a long derivation reflects
greater allotted effort or genuine difficulty, or whether the current branch is
becoming unproductive. Figure~\ref{fig:teaser}a makes this gap concrete: two
rollouts on the same problem emit the same local $15$-token span, yet a semantic
readout already separates them along named reasoning dimensions. Their
continuations then diverge, and only one succeeds. For test-time selection and
compute allocation, such distinctions matter earlier than the final answer can
reveal them.

The Jacobian lens maps an intermediate hidden state onto vocabulary-aligned
directions that represent what the model could later put into words
\citep{gurnee2026verbalizable}. Reading a state against them yields \jspace{}. A high reading for a word such as \emph{constraint} does not mean
that the model has emitted it, but that the current state supports a verbalizable
representation of the concept. \jspace{} can therefore expose considerations that remain
silent in the generated trace. Its raw form, however, carries
one coordinate per vocabulary item, which makes it too large and too redundant for
trajectory monitoring.

\begin{figure*}[t]
\centering
\includegraphics[width=\textwidth]{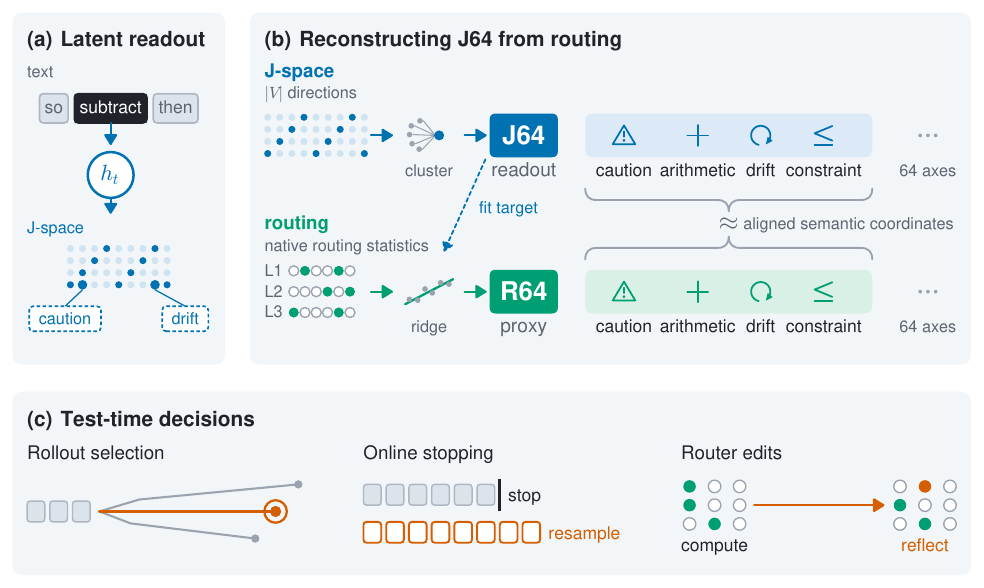}
\caption{\textbf{The instrument: a named readout, a routing proxy, three test-time
decisions.}
\textbf{(a)}~At a single token position we read the hidden state $h_t$ against
vocabulary-aligned directions, bringing out concepts the state supports whether or not
the rollout ever writes them; two of the high readings are named in dashed callouts, and
the rollout wrote neither word. A high reading for
\emph{caution} does not mean the model emitted the word.
\textbf{(b)}~Those directions are merged into the $64$ named axes of the \jsf{} frame,
built once per model and named only after construction (\S\ref{sec:rq1}). The same
reading is reachable in two ways, at very different cost: from hidden states, which means
replaying the rollout through the model, or from native expert routing, which generation
already emits. The braces mark the two routes as reaching the same $64$ coordinates, not
two readouts that happen to resemble each other: a ridge map is \emph{fitted} once, on
\jsf{} as its target, to reproduce \jsf{} from routing without outcome labels, and
recovers it at a median
per-axis agreement of $r=0.69$--$0.86$ (\S\ref{sec:rq2}), which is what the
$\approx$ records.
\textbf{(c)}~Routing statistics are emitted by generation itself, so reading R64 costs one
ridge projection rather than a second forward pass. R64 is therefore the
only copy available at serve time, and the one that acts: \emph{rollout selection}
among completed rollouts (\S\ref{sec:rq3}), \emph{online stopping} on a rolling window
during generation (\S\ref{sec:rq4}), and \emph{router edits} that move
a named axis to test the readout's causal semantics (\S\ref{sec:rq5}).}
\label{fig:pipeline}
\end{figure*}

We turn this vocabulary-scale readout into \jsf{}, a compact semantic frame
constructed without outcome, effort, or difficulty labels. From the model's own
reasoning states, we identify vocabulary directions that receive consistently
high \jspace{} readings, merge near-duplicates into $64$ semantic families, and
use each family's weighted mean direction as one axis. Axis names are assigned
only after construction and do not affect the readout. \jsf{} maps each hidden
state to a $64$-value reasoning dashboard, with readable axes such as \emph{caution},
\emph{arithmetic}, \emph{constraint}, and \emph{optionality}
(Figure~\ref{fig:pipeline}b). Window and
trajectory averages describe how the process evolves. \jsf{} is therefore a
coordinate system for latent reasoning state, rather than a correctness
classifier or a list of predicted tokens. Its axes stay active even when their
anchor words are absent from the nearby trace. They separate \emph{inference
posture} from \emph{problem-induced strain}, and they add $0.096$--$0.135$ held-out
AUC beyond a token-occupancy baseline that is aggregated in the same way.

\jsf{} makes latent state readable, but it requires hidden-state access. MoE
inference already emits expert assignments and gate weights at every token,
although raw routing carries no obvious meaning on its own. We therefore learn R64,
reconstructing the same $64$ semantic coordinates from native routing statistics
(Figure~\ref{fig:teaser}c).
Across three models and two families, R64 reaches median per-axis correlations
of $0.69$--$0.86$ and preserves $95$--$100\%$ of \jsf{}'s predictive gain on
gpt-oss-20b. R64 thus turns native routing into low-overhead, semantically grounded
reasoning-state telemetry.

Our contributions are threefold. First, \jsf{} provides a compact, readable
account of process state left implicit by the trace. Second, R64 reconstructs
that state from native MoE routing as a low-overhead deployment form. Third, this telemetry
acts at test time: it improves completed-rollout selection, majority-vote weighting in
seven of eight settings, and online stop-and-resample decisions, and router edits aimed
at the named mechanism change reasoning in the direction the readout predicts.

\section{Related Work}

\paragraph{Reading internal states.}
Vocabulary-aligned lenses and linear probes show that intermediate activations
can expose information that model outputs do not directly show
\citep{nostalgebraist2020logitlens,belrose2023tuned,azaria2023internal,%
marks2023geometry,zou2023representation,lindsey2025introspection}. Most closely,
\citet{gurnee2026verbalizable} introduce the Jacobian lens and identify
\jspace{} as a vocabulary-aligned set of verbalizable representations that can
carry silent intermediate reasoning. We use this readout operationally rather
than testing its global-workspace properties. Our contribution is to distill its
vocabulary-scale output into a compact trajectory-level frame, recover that
frame from native MoE routing, and use the resulting telemetry for test-time
selection and control.

\paragraph{Test-time compute and selection.}
Repeated sampling widens the reachable solution set
\citep{brown2024monkeys,snell2024scaling}. Majority voting aggregates the resulting
candidates \citep{wang2023selfconsistency}, and process reward models score them from
the text \citep{lightman2023verify}. A separate line lengthens reasoning without
touching the weights \citep{muennighoff2025s1}. Text-only self-correction remains hard
\citep{huang2024cannot,chen2024overthinking}. Our selector differs from these in
what it reads: internal state, or its routing-only proxy, instead of the trace. The
same telemetry also supports stopping and resampling \emph{during} generation, at a
point where answer-level consensus does not yet exist.

\paragraph{MoE routing.}
Routing is normally studied for capacity, load balancing and expert
specialization \citep{shazeer2017moe,fedus2022switch}, while behavioral control
is pursued by writing directions into the residual stream
\citep{turner2023activation,li2023iti}. Closest to our online experiments, DeepConf
\citep{fu2025deepconf} filters or early-stops reasoning traces on token-level
confidence: it needs no training, but reads only the output distribution. We use it
as the confidence baseline throughout, and treat routing instead as a semantic
sensor. After a one-time calibration against the lens, the usage spectrum
reconstructs an interpretable readout and inherits its downstream value. That
turns a load-balancing by-product into deployable telemetry.

\section{Method and Protocol}
\label{sec:method}
Figure~\ref{fig:pipeline} summarizes the instrument specified in this section: a
vocabulary-aligned readout of the hidden state, a reconstruction of it from native expert
routing, and the temporal resolutions at which the reconstruction is then read.

\paragraph{Models and data.}
Our primary model is gpt-oss-20b \citep{openai2025gptoss}, a mixture-of-experts
reasoning model whose discrete inference-effort setting (Low / Medium / High)
lengthens reasoning traces without changing weights. Replication runs on
gpt-oss-120b for scale and on Qwen3-30B-A3B (Thinking and Instruct) for a second
family. The dataset is competition mathematics: AIME-24, AIME-25, HMMT-25 and BRUMO-25 with
$30$ questions each, sampled at up to $64$ rollouts per question per setting. Architecture details, lens layer ranges and bank sizes are given in
Appendix~\ref{app:frame} and~\ref{app:r64}. The research questions act at
different points: RQ3 chooses among completed rollouts, RQ4 scores
$256$-token windows to decide whether to stop and resample, and RQ5 edits routing
during live generation. RQ1 and RQ4 count a run as successful only when it
naturally terminates with the correct answer, whereas RQ3 scores any extractable
answer (Appendix~\ref{app:labels}).

\paragraph{The \jsf{} readout.}
A state is described by $64$ concept readings, and a trajectory
$\tau=(y_1,\dots,y_T)$ by their mean,
$\phi_{\mathrm{J}}(\tau)=\frac{1}{T}\sum_t \jsf{}(h_t)$. In implementation the lens
is fitted per model and read at one layer. The candidate directions are seeded either
by sparse-coding mass or by lens-decode frequency, whichever passes two
construction-pool diagnostics for that model; every later construction step is
identical under both. The reading itself is
$\jsf{}(h)=A^{+}(h-\mu)$, where the frame is $A=[a_1\,\cdots\,a_{64}]$ and $A^{+}$ is
a pseudo-inverse rather than a transpose, because the axes are not
orthogonal. Algorithm~1 in Appendix~\ref{app:frame} gives the full construction: the
two seeders and how each model's seeder was chosen, the per-model layers, the units in
which readings are reported, and the axis names. The frame's axes are of two kinds:
some name a reasoning concept, and the rest mark what kind of text the model is
currently producing, such as digits, non-English tokens, or a change in writing
style. Appendix~\ref{app:rq1} lists every axis and how its name was assigned. On the
readability audit reported there, $51$ of the $64$ axes are highly readable and $85\%$ of the axes'
nearest-neighbor vocabulary is content words, against $32/64$ and $74\%$ for a
capacity-matched PCA basis of the same states.

\paragraph{The R64 proxy.}
We summarize three MoE layers per model as a usage spectrum: per expert, the gate
weight accumulated over the model's own routed assignments and normalized by the run's
token count, plus one gate entropy per layer. That is $99$ features for gpt-oss-20b and
$387$ for the two $128$-expert models. On that spectrum we learn a ridge map
$\mathrm{R64}(\tau)=g_\theta(\mathrm{routing}(\tau))$ onto the $64$ \jsf{} axes, using
no outcome labels: the map is calibrated against \jsf{} alone.
RQ2 measures reconstruction fidelity with question-held-out folds over a setting's
whole pool. The deployment experiments of RQ3 instead fit the map on the source
benchmark alone and freeze it before it sees any target.

\paragraph{What the proxy saves.}
The frame already contains the lens, so the \jsf{} projection is one $64\times d$ matrix
product. The cost is in obtaining the state it reads. Our generation stack emits expert
assignments and gate weights as a by-product of routing, but not activations, so a
\jsf{} reading means replaying the trajectory through the model in a second forward
pass. R64 reads what generation already produced and adds feature aggregation and one
light regression head. Fitting that map is a one-time calibration, and it is the one
step that does need the lens, since it requires trajectories with paired routing and
\jsf{} readouts. We fit it once per model and effort setting for RQ2, and once per
source benchmark within each transfer fold for RQ3; the map is frozen thereafter, and
deployment reads routing only. Measured on
gpt-oss-20b, decoding runs at $38.4$ ms per token. Capturing gate weights adds
$0.07$ ms and a \jsf{}-projection hook $0.09$ ms, both within run-to-run noise. The
sparse coding that seeds the frame costs $43$ ms per state and is paid once, at
construction.

Routing is read at two temporal resolutions, trajectory-aggregated in RQ2--RQ3 and
rolling-window in RQ4. The two share no parameters: the prefix failure score is a
separate head, not the trajectory-mean R64 vector. Three routing-level quantities
recur below and are distinct. R64 predicts the readout from the full spectrum, and the
module analysis of RQ2 describes how routing is organized. The intervention targets of
RQ5 are identified from \jsf{} geometry or outcome-linked usage, not from R64's
largest coefficients.

\paragraph{Statistical protocol.}
The statistical unit is the \emph{question}: model selection uses question-grouped
folds, and confidence intervals are question-level cluster bootstraps, with a
star (\sig{}) marking an interval that excludes zero. Cross-benchmark results use
frozen transfer: every fitted component is trained on one benchmark, frozen,
then tested on the other three. Supervision differs by component. The \jsf{} frame
uses no labels of any kind. The selection and online-control heads are trained with
correctness or failure labels on source questions, and the posture and strain
coordinates of RQ1 are selected on half the problems and confirmed on the other half.
On the $96$ evaluation questions that contribute no construction state, the principal
RQ1 diagnostics and \jsf{}'s selection advantage persist in the five gpt-oss-20b and
Qwen settings tested (Appendices~\ref{app:rq1} and~\ref{app:rq3proto}). All
labels use a verified answer parser, and every technical term is defined in
Appendix~\ref{app:glossary}.

\section{Building an Interpretable Routing Readout}
\label{sec:part1}

This part builds the instrument and asks two questions of it. The first is whether
\jsf{} captures readable, outcome-relevant process state beyond a summary of the
emitted trace built the same way, over the whole rollout and without token order
(RQ1). The second is whether that state can be read cheaply from the model's own
routing, and located within it (RQ2).

\subsection{RQ1: \jsf{} Reveals Process State Beyond the Trace}
\label{sec:rq1}

\begin{figure*}[t]
\centering
\includegraphics[width=\textwidth]{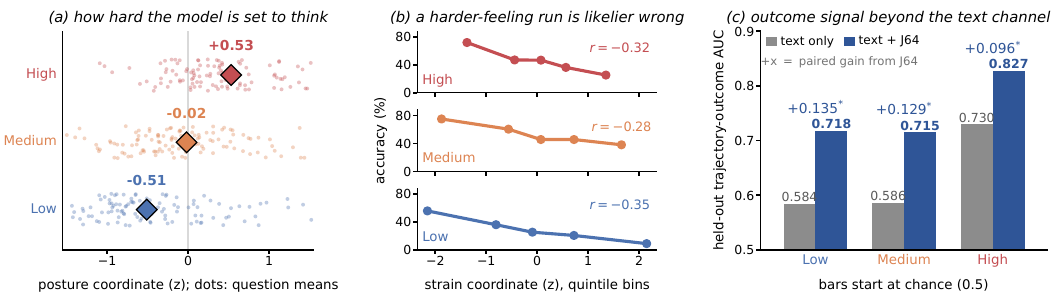}
\caption{\textbf{\jsf{} exposes readable latent process state and outcome signal beyond
the matched text channel.}
\textbf{(a)}~The posture coordinate separates the three effort settings. Dots are
question means.
\textbf{(b)}~Pooled \emph{rollout}-level accuracy across quintiles of the strain coordinate,
one row per effort with High at the top and Low at the bottom, as in panel~(a). Each
row's $r$ is the rollout-level correlation with correctness: $-0.32$ at High, $-0.28$
at Medium and $-0.35$ at Low. It is \emph{not} the problem-level correlation with
difficulty, which is $+0.41/+0.46/+0.41$ at Low/Medium/High.
\textbf{(c)}~Adding \jsf{} to the matched text representation raises held-out
trajectory-outcome AUC at every effort. The bars are the two channels' AUC; the
$+$ value above each pair is the \emph{paired} increment, estimated per question
and question-clustered, with \sig{} marking an interval that excludes zero. It is
therefore not exactly the difference of the two rounded bar labels.}
\label{fig:coords}
\end{figure*}

RQ1 asks whether a reasoning trace leaves out process state that remains readable
inside the model. Figure~\ref{fig:teaser}a makes that gap concrete with a pair of
rollouts matched on the emitted text. We establish the claim in three steps:
\jsf{} is not a direct lexical echo; its readings organize into interpretable
process coordinates; and it adds outcome-relevant information beyond a
representation of the emitted rollout that is aggregated in the same way.

\paragraph{The readout is not a lexical echo.}
We call an axis's nearest vocabulary neighbors its token family. Consider the
positions where an axis reads above two standard deviations. On average $99.6\%$ of
those positions have no member of that axis's own token family within eight tokens,
and for $50$ of the $64$ axes the figure is exactly $100\%$. The family words themselves appear
at most $0.7$ times per thousand tokens. When a family word does appear the reading rises sharply, by up to $+3.9$z, and
it does not rise at other axes' family words, where the mean is $z=-0.07$. The frame is
lexically grounded where the text allows a check, and it stays active where the text is
silent. That combination is what a readout of internal state can offer over a
representation of the emitted tokens.

\paragraph{Two readable process coordinates.}
Having established that \jsf{} is not a direct copy of the emitted words, we next
ask what latent process distinctions its axes organize. The frame's structure
appears in two fixed contrasts of the form $z=w^\top\phi_{\mathrm{J}}(\tau)$. In each,
$w$ weights one group of axes positively and a second group negatively, so $z$ is high
when the first group reads high relative to the second. The \emph{posture} coordinate contrasts a three-axis case-splitting bundle
against a six-axis arithmetic core. That bundle covers optional cases,
case-by-case transitions and one-by-one processing. It steps monotonically with the effort dial at
$-0.51/\!-\!0.02/\!+\!0.53$z, consistently across all four benchmarks
(Figure~\ref{fig:coords}a), and carries no within-problem outcome signal, so it reads
how long the model has been set to think. The \emph{strain} coordinate contrasts the
problem-perception axis against the constraint-requirement axis, $w=e_{30}-e_{1}$,
where $e_k$ selects axis $k$ of the frame. Readings are standardized over the analysis
sample before either contrast is taken, so $z$ is in units of a standard deviation.
Appendix~\ref{app:frame} gives the axis sets, the group-size normalization that
\emph{posture} uses, and the half-split that selected them. Strain is
effort-blind: its tier means lie within $0.04$z of zero. What it tracks instead is problem difficulty, at
$r=0.41/0.46/0.41$; on $60$ held-out problems the same correlations are
$0.28$\sig{}/$0.35$\sig{}/$0.28$\sig{}. Rollout accuracy falls monotonically
across its quintiles at every effort (Figure~\ref{fig:coords}b). The two coordinates
are near-orthogonal. How hard the model has been asked to work and how hard the
problem itself is are therefore two separate coordinates of the readout, not one
shared axis of general activation.

\paragraph{The outcome increment over matched text.}
Both channels summarize the rollout the same way: each collapses all $T$ positions
into a single vector without using token order. Only the source differs, so a
difference in AUC is attributable to what is read rather than to how the sequence is
summarized. The text channel is the occupancy of emitted tokens over a
$3{,}000$-token vocabulary, compressed and scaled in the way count features usually
are. Let $c(\tau)\in\mathbb{R}^{3000}$ hold the token counts of the rollout; then
$\phi_{\text{text}}(\tau)=\log(1{+}c(\tau))/\lVert\log(1{+}c(\tau))\rVert_2$. The
internal channel is $\phi_{\mathrm{J}}(\tau)$ of \S\ref{sec:method}. The two also share
question splits, classifier family and regularization, and the text side is handed
$47\times$ more features than \jsf{}. AUC is a
within-problem paired comparison pooled over problems (Appendix~\ref{app:labels}), so
problem identity does not enter the score. Adding \jsf{} to the matched text model
raises it at every effort setting, from
$0.584$ to $0.718$ at Low ($+0.135$\sig{}), from $0.586$ to $0.715$ at Medium
($+0.129$\sig{}) and from $0.730$ to $0.827$ at High ($+0.096$\sig{}). At $120$b scale
the readout alone remains far stronger than the matched text model
($0.794/0.761/0.865$ against $0.585/0.613/0.722$). We report the two channels
separately at that scale because a stacked fit there is dominated by the
$3{,}000$-feature text block (Appendix~\ref{app:scale}). Instrument
details and additional analyses are in Appendix~\ref{app:rq1}.

\jsf{} therefore exposes a compact, semantically readable process state that is not
directly shown by the emitted trace and that carries outcome information beyond
the matched text channel. The remaining question is whether the same hidden state
can be recovered from a signal that generation already emits, without the second
forward pass the lens requires.

\subsection{RQ2: Routing Reads Out and Localizes \jsf{}}
\label{sec:rq2}

\begin{figure}[t]
\centering
\includegraphics[width=\columnwidth]{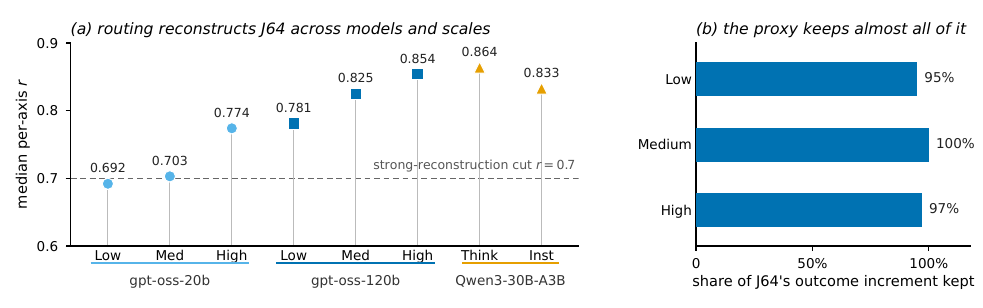}
\caption{\textbf{Routing is a faithful, rollout-specific proxy for \jsf{}.}
\textbf{(a)}~Median per-axis held-out reconstruction in the eight model--effort settings,
each read through its own frame. The dashed line is the strong-reconstruction cut at
$r{=}0.7$, which $28$ to $59$ of a setting's $64$ axes clear (shuffled routing:
$r\approx0$). Marker shape and color give the model
family. \textbf{(b)}~The share of
\jsf{}'s outcome increment over matched token occupancy that the routing-only
reconstruction keeps.}
\label{fig:proxy}
\end{figure}

\begin{figure}[t]
\centering
\includegraphics[width=\columnwidth]{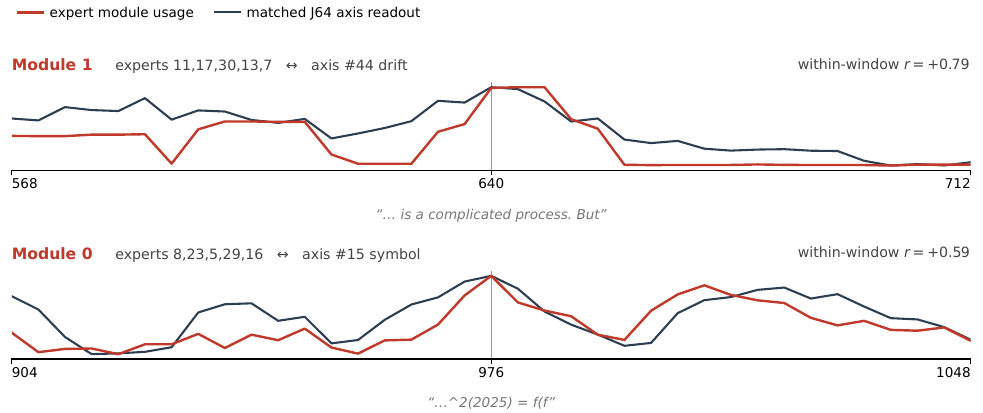}
\caption{\textbf{Routing indicates \emph{which stage} the reasoning is in.} Two of the eight
expert modules and the \jsf{} axes they lock onto, along one High-effort trace:
module usage in red, matched axis readout in grey, both curves min--max scaled within
the window. Each panel is a $\pm72$-token window centered where the two curves are
jointly highest. Module~1 tracks the drift-and-reconsideration axis and peaks on the
paragraph break of ``\dots{}a complicated process. But''; Module~0 tracks the symbol
axis and peaks inside a symbolic expression. All eight modules,
their enriched vocabulary and the matched axes are in Appendix~\ref{app:rq2}.}
\label{fig:modules}
\end{figure}

RQ2 asks whether the same readable state can be recovered from routing records
already produced during generation, without replaying hidden activations.

\paragraph{Fidelity.}
A ridge model maps the routing usage spectrum onto the $64$ \jsf{} axes with
question-held-out folds. Integrated over a run, median per-axis reconstruction is
$0.692$--$0.864$ across the eight model--effort settings of
Table~\ref{tab:bo64transfer}, and it rises with model scale
(Figure~\ref{fig:proxy}a; per-axis distributions in Figure~\ref{fig:peraxis}). The
two Qwen variants reconstruct at $0.864$ and $0.833$. Reconstruction from a
\emph{single} token's routing still reaches $r=0.42$--$0.48$. Three robustness checks
support this. Shuffling routing against the readouts collapses every model--effort setting to
$r\approx0$; a single MoE layer suffices; and,
\emph{for the instantaneous token-level map}, transferring across effort settings
costs at most $0.04$ (Figure~\ref{fig:transfer}). The correspondence is therefore a structural property of MoE
reasoning rather than an artifact of one model: each family is read through its
own model-specific frame under identical folds and regularization. The $120$b
replication is in Appendix~\ref{app:scale}.

\paragraph{Specificity.}
Routing could reconstruct \jsf{} merely because both echo the prompt, the emitted
words or the trace length. We residualize \jsf{} against a $300$-dimensional PCA of full-trace token occupancy
plus quadratic log-length. Routing still explains additional held-out $R^2$ on that
residual: $0.082$ $[0.072,0.093]$ at Low effort and $0.085$ $[0.072,0.098]$ at
Medium. When routing traces are permuted among sibling rollouts of the same question,
the same quantity falls to $0.015$ and $0.011$, five to eight times smaller. The correspondence
therefore tracks the individual rollout, not the question: it disappears when the
pairing between a rollout and its own routing is broken. It also replicates at
$120$b scale, where reconstruction after removing each question's mean still
reaches $r=0.72/0.76/0.80$.

\paragraph{Utility.}
The proxy recovers the predictive content as well as the geometry. Against the same
matched text baseline R64 adds $+0.128$\sig{}/$+0.129$\sig{}/$+0.093$\sig{} outcome AUC
at Low/Medium/High, which is $95$--$100\%$ of \jsf{}'s increment
(Figure~\ref{fig:proxy}b). A plain ridge therefore preserves nearly all of \jsf{}'s
predictive content, while routing needs no additional forward pass and the lens is
queried only once, at calibration.

\paragraph{What the proxy is reading.}
The routing features the ridge reads inherit interpretable semantics from \jsf{}.
We factorize per-token expert usage at $347$k positions into eight nonnegative modules
and correlate each module against the mean-removed axis readouts. Every module locks
onto an axis that RQ1 had \emph{already named} from the lens vocabulary, before any
routing analysis, so each arrives with a semantic hypothesis attached rather than a
label we chose. One module tracks the drift-and-reconsideration axis and is enriched on \emph{But},
\emph{Now}, \emph{Wait} and \emph{Alternatively}, the vocabulary of abandoning the
current line. A second tracks symbolic algebra and peaks inside mathematical
expressions; a third tracks optionality and is enriched on the modal verbs
\emph{might}, \emph{will}, \emph{cannot} and \emph{should}. Shuffling the readouts
collapses the largest correlation to $0.005$.

The correspondence holds token by token, not only in aggregate.
Figure~\ref{fig:modules} follows the first two modules along a single High-effort
trace. Expert usage and the matched axis rise and fall together across a
$\pm72$-token window. Within that window the drift module correlates with its axis at
$r={+}0.79$ and the symbol module with its axis at $r={+}0.59$. The position-wide
values are $+0.59$ and $+0.48$. Each peak also lands where the named
state should be active: the drift module peaks on a paragraph break before
\emph{But}, and the symbol module inside an
expression. Table~\ref{tab:modules} names all eight modules and the axis each one
matches. Routing therefore does not merely predict a trajectory-mean readout;
it follows the process state from token to token. The full eight-module
analysis, the enrichment procedure and the two limits on this reading
are in Appendix~\ref{app:rq2}.

Reconstruction and module structure give routing two complementary roles. R64 makes
the \jsf{} state observable from native expert usage, with semantics inherited from
the frame, and it supports the selection and online policies of RQ3--RQ4. The module
analysis localizes named process states to compact expert groups, and provides the
mechanism hypotheses that RQ5 tests: there, the same
decomposition places the causally effective experts in compact layer-specific modules.

\section{Acting on the Readout at Test Time}
\label{sec:part2}

This part uses the routing signal in two ways. As telemetry, it chooses among
finished trajectories and reallocates compute during generation (RQ3--RQ4). As a
mechanism, it supports targeted intervention on the computation that the readout
names (RQ5).

\subsection{RQ3: Completed-Rollout Selection and Voting}
\label{sec:rq3}

\begin{table*}[!t]
\centering
\footnotesize
\setlength{\tabcolsep}{1.3pt}
\resizebox{\textwidth}{!}{%
\begin{tabular}{l cccc cccc cc}
\toprule
 & \multicolumn{4}{c}{gpt-oss-20b} & \multicolumn{4}{c}{gpt-oss-120b} & \multicolumn{2}{c}{Qwen3-30B-A3B} \\
\cmidrule(lr){2-5}\cmidrule(lr){6-9}\cmidrule(lr){10-11}
 & Low & Med & High & \emph{avg} & Low & Med & High & \emph{avg} & Think & Inst \\
\midrule
Avg@64 (random pick) & 35.1 & 66.5 & 70.9 & 57.5 & 48.8 & 73.3 & 83.1 & 68.4 & 78.8 & 60.9 \\
\midrule
\multicolumn{11}{l}{\emph{single-trace selectors --- frozen transfer, averaged over all four training benchmarks}} \\
J64 & \textbf{41.9}\gp{6.9} & 68.1\gp{1.6} & \textbf{77.5}\gp{6.6} & \textbf{62.5}\gp{5.0} & \textbf{54.4}\gp{5.6} & \textbf{76.9}\gp{3.7} & 88.6\gp{5.5} & \textbf{73.3}\gp{4.9} & 85.3\gp{6.5} & 65.6\gp{4.7} \\
R64 (routing) & 35.8\gp{0.8} & \textbf{68.3}\gp{1.9} & 76.1\gp{5.2} & 60.1\gp{2.6} & 51.1\gp{2.3} & 75.6\gp{2.3} & 86.1\gp{3.0} & 70.9\gp{2.5} & \textbf{87.2}\gp{8.4} & 64.7\gp{3.8} \\
token occupancy & 35.6\gp{0.5} & 64.2\gm{2.3} & 74.7\gp{3.8} & 58.2\gp{0.7} & 53.9\gp{5.1} & 75.6\gp{2.3} & 86.9\gp{3.8} & 72.1\gp{3.7} & \textbf{87.2}\gp{8.4} & 65.6\gp{4.7} \\
length & 20.8\gm{14.2} & 60.0\gm{6.5} & 73.3\gp{2.4} & 51.4\gm{6.1} & 35.0\gm{13.8} & 65.0\gm{8.3} & \textbf{89.2}\gp{6.1} & 63.1\gm{5.3} & 84.2\gp{5.4} & 61.7\gp{0.8} \\
DeepConf & 41.7\gp{6.6} & 62.5\gm{4.0} & 60.8\gm{10.1} & 55.0\gm{2.5} & 51.7\gp{2.9} & 76.7\gp{3.4} & 82.5\gm{0.6} & 70.3\gp{1.9} & 80.8\gp{2.0} & \textbf{65.8}\gp{4.9} \\
\midrule
\multicolumn{11}{l}{\emph{consensus family --- vote weights from the same frozen selectors}} \\
majority vote & 46.7\gp{11.6} & 83.3\gp{16.9} & 85.0\gp{14.1} & 71.7\gp{14.2} & 58.3\gp{9.5} & 84.2\gp{10.9} & 95.0\gp{11.9} & 79.2\gp{10.8} & 90.8\gp{12.0} & 70.8\gp{9.9} \\
vote $+$ DeepConf & \textbf{50.0}\gp{14.9} & 80.8\gp{14.3} & 85.0\gp{14.1} & 71.9\gp{14.4} & 58.3\gp{9.5} & 85.0\gp{11.7} & \textbf{95.8}\gp{12.7} & 79.7\gp{11.3} & 90.0\gp{11.2} & 71.7\gp{10.8} \\
vote $+$ J64 & 48.1\gp{13.0} & \textbf{84.4}\gp{18.0} & 83.9\gp{13.0} & \textbf{72.1}\gp{14.6} & 58.1\gp{9.3} & 86.7\gp{13.4} & 95.0\gp{11.9} & 79.9\gp{11.5} & 89.4\gp{10.6} & 71.4\gp{10.5} \\
vote $+$ R64 & 47.8\gp{12.7} & 83.1\gp{16.6} & \textbf{85.3}\gp{14.4} & \textbf{72.1}\gp{14.6} & \textbf{59.2}\gp{10.4} & \textbf{87.8}\gp{14.5} & 95.6\gp{12.5} & \textbf{80.8}\gp{12.4} & \textbf{91.1}\gp{12.3} & \textbf{73.1}\gp{12.2} \\
\bottomrule
\end{tabular}%
}
\caption{\textbf{Cross-benchmark frozen
transfer, best-of-64 (\%), across eight deployment settings.} Every learned method,
including R64's reconstruction ridge and the token-occupancy vocabulary, is trained on
one benchmark and applied unchanged to the other three. Each cell averages
that setting's $12$ off-diagonal transfer pairs (deterministic single pick, all $30$
questions per test set). Branches are scored by extracted answer (Appendix~\ref{app:labels}).
Parenthesized values are the gain over the Avg@64 reference of
the same column, i.e.\ over a random pick. The \emph{avg} columns average a model's
three effort settings. Bold $=$ best per column within each block.}
\label{tab:bo64transfer}
\end{table*}

\begin{table}[t]
\centering
\small
\setlength{\tabcolsep}{1.8pt}
\begin{tabular}{lcccc}
\toprule
 & oss-20b & oss-120b & Qwen & \textbf{pooled} \\
failed questions & 102 & 75 & 46 & 223 \\
\midrule
random pick & 8.5 & 10.6 & 5.3 & 8.7 \\
J64 & \textbf{14.3}\gp{5.8} & \textbf{23.2}\gp{12.6} & \textbf{19.2}\gp{13.9} & \textbf{17.0}\gp{8.3}\sig{} \\
R64 (routing) & 12.3\gp{3.8} & 20.0\gp{9.3} & 16.1\gp{10.8} & 14.1\gp{5.3}\sig{} \\
token occupancy & 8.1\gm{0.4} & 14.6\gp{3.9} & 15.1\gp{9.8} & 12.3\gp{3.5}\sig{} \\
length & 3.3\gm{5.2} & 14.6\gp{4.0} & 11.7\gp{6.4} & 5.4\gm{3.4}\sig{} \\
DeepConf & 9.4\gp{0.8} & 20.5\gp{9.9} & 11.7\gp{6.4} & 13.0\gp{4.3}\sig{} \\
\bottomrule
\end{tabular}
\caption{\textbf{Rescue on the questions where plain majority voting scores zero (\%), by
model.} Protocol identical to Table~\ref{tab:bo64transfer}, restricted to the failed
questions. Model columns average that model's effort settings, and \textbf{pooled} is
the rate over all $223$. Question-paired clustered CIs: \jsf{}
$+8.3$\sig{}$[+5.3,+11.3]$, R64 $+5.3$\sig{}$[+2.3,+8.5]$, length
$-3.4$\sig{}$[-6.4,-0.1]$. Per-setting detail is in Appendix~\ref{app:rq3}.}
\label{tab:rescue}
\end{table}

RQ3 tests the offline use case: after $64$ sibling rollouts are available, can the
readout choose a stronger branch or improve their aggregation?
Table~\ref{tab:bo64transfer} is the deployment test under
frozen cross-benchmark transfer: every fitted component is trained on a single
source benchmark, frozen, and applied unchanged to the other three. Every
number averages over all four choices of training benchmark. Two components are
shared globally. The first is the unlabeled \jsf{}
frame, built once per model from a calibration pool spanning the four benchmarks
(\S\ref{sec:method}). The second is a single regularization strength $C{=}3$, chosen once by
question-held-out cross-validation inside one benchmark.
Appendix~\ref{app:rq3proto} reports a stricter source-only control. Baselines are
trace length, a matched token-occupancy selector, random choice, and a training-free
confidence method \citep[DeepConf;][]{fu2025deepconf}. The majority-vote family forms
a separate class, because it consults all $N$ branches at once.

\paragraph{Single-branch selection.}
\jsf{} and R64 both improve single-branch selection over a random candidate, in every
model group. \jsf{} is the most consistent single-trace method, with the highest
effort-averaged accuracy at both gpt-oss scales, $+5.0$ and $+4.9$ points over a random
pick, and it leads four of the eight individual settings. R64 stays positive in all four
model groups, at $+2.6$, $+2.5$, $+8.4$ and $+3.8$ points, so the routing form of the
readout is deployable here without reading hidden states. Trace length illustrates
what a single-signal selector buys: it wins in the settings where truncation is the
dominant failure mode and collapses in the settings where it is not
(Appendix~\ref{app:rq3}).

\paragraph{A second task domain.}
We next test the same selection objective on GPQA \citep{rein2023gpqa}, $198$ graduate-level science
questions at $64$ rollouts per setting, using question-held-out,
question-residualized folds where no cross-benchmark transfer is possible
(Table~\ref{tab:gpqasel} in Appendix~\ref{app:rq3}). Pooled over all $990$ question--setting units, only the two internal channels clear
zero, \jsf{} at $+4.6$\sig{} and R64 at $+4.3$\sig{}, while token occupancy, length
and DeepConf show no consistent positive gain.

\paragraph{Vote weighting and rescue.}
Consensus remains the strongest aggregate rule, and routing is most useful as a weight on
it rather than as a replacement. We weight each branch's vote by the frozen
selector's predicted correctness probability. This improves plain majority voting in
seven of the eight settings and in all four model-group averages
($+0.4$/$+1.6$/$+0.3$/$+2.3$ for the routing-weighted variant). The sharpest test
comes from the $223$ questions, across the eight settings, on which plain majority
voting scores zero (Table~\ref{tab:rescue}). We run the identical frozen-transfer
protocol on those questions. \jsf{} selects a correct branch on $17.0\%$ of them and
R64 on $14.1\%$, against $8.7\%$ for a random pick, and the ordering is the same in
all three models. That
slice is picked with hindsight; it shows that the selectors and majority voting fail
on different questions. Appendix~\ref{app:rq3} reports the pool composition and the
recoverable ceiling.

Completed-rollout telemetry improves the decision, but it cannot recover compute already
spent on weak branches.

\subsection{RQ4: Online Compute Allocation}
\label{sec:rq4}

\begin{table}[t]
\centering
\small
\setlength{\tabcolsep}{2.6pt}
\begin{tabular}{lcccc}
\toprule
Target cost & \jsf{} & R64 & DeepConf & permSib \\
\midrule
\multicolumn{5}{l}{\emph{gpt-oss-20b High} --- one uninterrupted rollout: $56.8$} \\
$1.2\times$ & $58.3$\,/\,$1.21$ & $\mathbf{59.2}$\,/\,$1.21$ & $58.7$\,/\,$1.22$ & $57.0$\,/\,$1.20$ \\
$1.4\times$ & $\mathbf{62.2}$\,/\,$1.42$ & $60.8$\,/\,$1.41$ & $60.0$\,/\,$1.44$ & $57.8$\,/\,$1.45$ \\
$1.6\times$ & $\mathbf{62.9}$\,/\,$1.59$ & $60.8$\,/\,$1.50$ & $59.8$\,/\,$1.54$ & $57.7$\,/\,$1.57$ \\
$2.0\times$ & $\mathbf{63.5}$\,/\,$1.65$ & $60.8$\,/\,$1.50$ & $59.4$\,/\,$1.76$ & $57.6$\,/\,$1.67$ \\
\midrule
\multicolumn{5}{l}{\emph{Qwen3-30B-A3B Thinking} --- one uninterrupted rollout: $78.1$} \\
$1.4\times$ & $\mathbf{79.6}$\,/\,$1.43$ & $79.5$\,/\,$1.43$ & $79.2$\,/\,$1.36$ & $78.6$\,/\,$1.42$ \\
$1.6\times$ & $\mathbf{80.0}$\,/\,$1.63$ & $79.9$\,/\,$1.65$ & $79.2$\,/\,$1.36$ & $78.5$\,/\,$1.56$ \\
$2.0\times$ & $\mathbf{80.1}$\,/\,$1.92$ & $\mathbf{80.1}$\,/\,$1.88$ & $79.2$\,/\,$1.36$ & $78.4$\,/\,$1.79$ \\
\bottomrule
\end{tabular}
\caption{\textbf{Cumulative stop-and-resample accuracy under the nested protocol.} Each cell
reports accuracy (\%)\,/\,\emph{realized} test cost, the latter as a multiple of one
uninterrupted rollout. Outcomes count delivered answers only, hence the reference
level sits below Table~\ref{tab:bo64transfer}'s random pick (Appendix~\ref{app:labels}).
Question-paired differences against \textbf{permSib} are in the text. Bold $=$ best
per row.}
\label{tab:cusum}
\end{table}

RQ4 tests the online use case: before a rollout finishes, can the readout identify
an unproductive attempt early enough to stop and resample?
Every $256$ generated tokens, a window-level head scores failure risk from \jsf{}, R64
or DeepConf. A CUSUM controller accumulates that evidence as
$S_t=\max(0,\,S_{t-1}+s_t-\kappa)$, where $\kappa$ is the median window score. It
stops an attempt once the statistic crosses a threshold $\tau$, and generation then
restarts from a fresh sample. After at most $K$ cuts, the next attempt runs to its
natural stop.
Decisions are causally masked, restarts are drawn from a fixed independent pool, and
tokens are charged exactly. Every signal runs inside this same controller, with
only the window score swapped. The primary control exchanges whole score sequences between sibling rollouts of the same
question, which removes branch identity while preserving the marginal, the
autocorrelation and everything the score knows about the question. Every constant is
in Appendix~\ref{app:rq4proto}.

\paragraph{Choosing the operating point without the test questions.}
A threshold that fires earlier spends less, so the methods cannot be compared at a
nominal budget. Nor can we simply pick each method's best configuration at a given
cost: that would let every method profit from the questions it is scored on. We
therefore nest the selection. In each of four outer question folds we build an
accuracy--cost frontier on the training questions alone. We then take the two
configurations just below and just above the target cost, and freeze a randomized
mixture of them, with the mixture weights set so that its expected training cost
equals the target (Appendix~\ref{app:rq4proto}). Table~\ref{tab:cusum} reports its
accuracy and realized cost on the held-out questions. Nothing is selected on the
questions being scored, so the absolute levels are not optimistically biased.

\paragraph{What the comparison shows.}
The controller works on branch-level information. \jsf{} beats the sibling-permuted
control at every target cost: by $+1.3$\sig{}, $+4.4$\sig{},
$+5.2$\sig{} and $+5.9$\sig{} points at $1.2$, $1.4$, $1.6$ and $2.0\times$ on
gpt-oss-20b High, and by $+1.1$\sig{}, $+1.5$\sig{} and $+1.7$\sig{} on Qwen. The gain
comes from knowing \emph{which} attempt will fail, not from the freedom to restart
or from the shape of the score. Routing carries most of that information on
its own: R64 clears the same control by $+2.2$ to $+3.2$\sig{} points on High and
$+0.9$ to $+1.7$\sig{} on Qwen. The deployable form of the readout therefore retains
most of the benefit, which matters because it is the form a serving system would
actually run. The direct \jsf{}--R64 contrast is in Appendix~\ref{app:rq4}. DeepConf
is the weakest of the three inside the same controller. It
clears the control on High, but \jsf{} leads it by $2.2$ to $4.2$\sig{} points at
$1.4$--$2\times$. It matches \jsf{} at $1.2\times$, and on Qwen it does not separate
from the control at all.
Branch-specific telemetry remains useful even under tight compute, where it is what
makes restarting worth the tokens it costs. Routing serves here as a
process sensor, reallocating compute by stopping and resampling.

\subsection{RQ5: From Semantic Readout to Routing Mechanism}
\label{sec:rq5}

RQ5 asks whether the named states correspond to routing mechanisms whose editing
changes reasoning in the predicted direction. We study two complementary cases.
The first starts from a \jsf{} diagnosis and ranks experts by write-vector
advantage; the second starts from an outcome-linked routing group and interprets
it through \jsf{}. We edit router logits only. Appendix~\ref{app:rq5} gives the full
targets, edit strengths and conditions.

\paragraph{Amplifying the diagnosed failure.}
Non-terminating trajectories load on the case-splitting bundle. Raising the logits
of experts that advance this bundle drives accuracy from $0.381$ to $0.000$
($-0.381$\sig{}) and sends $0.98$ of runs to the generation limit. A sham edit of
equal strength, aimed at unrelated experts, is equally damaging in accuracy, but only
the targeted condition shows the predicted behavior: sustained ``Case $n$''
enumeration, reaching $100$ headers in one run, versus none in $160$ sham runs. The
edit therefore changes the form of failure in the direction named by \jsf{}, beyond
generic disruption.

\paragraph{Suppressing the diagnosed stall.}
A compact expert group is overused during the middle fifth of incorrect
trajectories ($0.085$ versus $0.021$), and the expert-wise effects persist after
controlling for trace length. Read through \jsf{}, the group's layer-$20$ member
corresponds to a state that remains on the problem's requirements instead of executing
the operation they call for. The targets also occupy compact unsupervised routing
modules, so the group is one the model's own routing already
forms. Suppressing the group shortens generations by $45.9$
tokens\sig{}, whereas the equal-strength sham lengthens them by $47.6$\sig{}. On the
focal tetrahedron problem, the intervention replaces early numerical guessing with
an exact symbolic derivation and recovers the correct answer. Across both samples in
this second study, every treatment condition is at or above the do-nothing control,
and the sham is the only one below it (Appendix~\ref{app:rq5}).

Together, these edits link \jsf{} semantics to routing mechanisms whose
manipulation produces the named reasoning change.

\section{Limitations and Conclusion}

Our analysis centers on competition mathematics and gpt-oss-20b. Reconstruction
and selection replicate on gpt-oss-120b and Qwen3-30B-A3B, but each model's frame is
built independently, so its axes are not aligned with any other model's. Online
control is evaluated mainly by causally masked replay, and the interventions remain
small. The readout tracks process state, and before an answer is committed it does
not predict the eventual outcome; its activity on reflective vocabulary is also
largely shared with the emitted distribution. Sequence-aware text baselines, live
serving and broader domains are left to future work.

\jsf{} makes readable the process state left implicit by the trace; R64
reconstructs it from native routing without activation replay. The same telemetry
supports completed-rollout selection and online stop-and-resample control, while
edits aimed at the named mechanism provide causal case studies. \jsf{} supplies semantics,
routing supplies deployment, and test-time policies make the signal actionable.

\small
\bibliographystyle{plainnat}
\bibliography{references}
\normalsize

\appendix
\section{Supplementary Analyses}
\label{app:analyses}

This appendix is grouped by research question and follows the order of the main text.
Appendix~\ref{app:repro} then specifies every construction, protocol and constant the
main text uses.

\subsection{RQ1: The Readout and Where Its Increment Sits}
\label{app:rq1}

\paragraph{Instrument details.}
\emph{Naming scheme.} Each axis carries a name and a responsibility tier:
sixteen are \emph{earned} (at text positions where an axis's family words
appear, its reading rises sharply --- \emph{accumulate} $+3.93$, \emph{hover}
$+3.40$, geometry $+2.74$z); nineteen are \emph{lens-named} from a coherent
neighbor family whose words scarcely appear in the text (itself evidence the
axis is subverbal); three are \emph{renamed} because grounding rejects the
surface name (the effort-sorting ``Unique'' axis is really a title-register
marker); twenty-six are \emph{state markers} (digit, multilingual, and
register directions, read for position on the manifold rather than literally).
Validation: a family word lifts its own axis by $z=+1.30$ but other axes by
only $-0.07$; the caution axis tracks written checking at run-level $+0.45$;
and $99\%$ of an axis's high-activation positions carry no family word within
eight tokens. \emph{Compactness.} Correctness signal is compact (an eight-axis
panel transfers $85$--$95\%$ of the increment across efforts), but effort
classification and best-of-$N$ selection need the whole dashboard (an
eight-axis frame loses $9\pp$\sig{} on selection).

\paragraph{Frame-disjoint stability check.}
We repeat the principal RQ1 diagnostics on the $96$ questions that contribute no
states to frame construction. The qualitative findings persist: effort
classification reaches $0.938$ AUC against $0.935$ on the full pool; difficulty AUC
is $0.694/0.794/0.668$ against $0.659/0.779/0.744$ at Low/Medium/High; and the
strain--difficulty correlations are $0.434/0.500/0.445$ against $0.410/0.460/0.405$.
The central frame-level effects are therefore not concentrated on the $24$
construction questions --- the High-effort difficulty estimate is lower on the
disjoint subset but stays well above chance, and the remaining audited effects
are comparable or stronger. Rerunning the Table~\ref{tab:bo64transfer} protocol on
the same $96$ questions, \jsf{}'s advantage over a random pick persists in all
five settings rerun, $+9.7/+1.9/+5.2$ on gpt-oss-20b Low/Medium/High and
$+9.2/+6.1$ on Qwen Thinking/Instruct. On the construction questions alone, effort classification reads
$0.894$ on $1{,}156$ runs, below both full-set values.

\paragraph{A capacity-matched PCA basis of the same states.}
The capacity control replaces the curated frame with PC64, the top $64$ principal
components of the same pooled states ($51.3\%$ of variance, orthonormal). The two
readouts span nearly the same subspace: the top five canonical correlations lie at
$0.986$--$0.999$ at every effort, $19$--$23$ of the $64$ directions correlate above
$0.9$, and selectors trained on either basis score the same rollouts alike
($r=0.74$--$0.88$). Used for selection under the same frozen evaluation, the
question-paired difference between the two is $-3.3$ $[-7.8,+1.1]$, $-2.2$
$[-6.7,+2.5]$ and $-1.7$ $[-4.0,+0.3]$ points at Low/Medium/High, and
concatenating the two readouts improves neither arm ($+0.6$ to $+2.8$ over \jsf{},
every interval crossing zero), so the second basis carries no additional
information. The small PC64 edge is an estimation property of an orthonormal basis
--- condition number $1$, against an effective rank of $33.8$ and a pseudo-inverse
readout for the non-orthogonal frame --- rather than additional signal.
What the principal components lack is names. Individually they are poorly aligned
with the frame (median $|\cos|=0.04$) and spend their dimensions on register and
punctuation, so data-driven directions do not land on reasoning concepts:
$0.85$ of \jsf{}'s axes' nearest-neighbor vocabulary is clean content words and
$51/64$ axes are highly readable, against $0.74$ and $32/64$ for PC64. The
analyses that carry this paper's mechanism content, the module--axis
correspondences of RQ2 and the target selection of RQ5, consume named
axes. The NMF-module selector of Appendix~\ref{app:rq2} fails for the
complementary reason: there the unsupervised step changes the subspace and loses
the predictive direction before any supervised step runs; here only the basis
changes and nothing is lost.

\subsection{Replication at $120$b Scale}
\label{app:scale}

\paragraph{RQ1--RQ2 at $120$b scale.}
On gpt-oss-120b (three effort settings, $7{,}680$ rollouts each) the
\emph{standalone} readout replicates and strengthens, while the stacked increment
does not (below): \jsf{} alone reads trajectory outcome at
$0.794/0.761/0.865$ versus $0.585/0.613/0.722$ for the matched full-trace text model,
and difficulty at $0.670/0.715/0.596$ versus $0.362/0.456/0.492$. Within-question
residual correctness --- the strict test that freezes the problem --- excludes
zero at all three efforts from the readout ($0.665$\sig{}/$0.672$\sig{}/%
$0.704$\sig{}) and from routing alone ($0.648/0.632/0.706$), and within-question
residual truncation reaches $0.935$ at High. First, the \emph{stacked} estimator (text $+$ \jsf{} in one $3{,}064$-%
dimensional model) adds nothing over text at this scale ($+0.004/+0.001/-0.001$,
CIs crossing zero) even though the readout alone is far stronger: with
$3{,}000$ occupancy features the regularized fit is dominated by the text block,
so at $120$b we report the single-arm duel rather than the stacked increment.
Second, the $120$b frame is extracted independently at that model's own layer,
so its axes are not aligned one-to-one with the $20$b axes; scale comparisons
are made at the level of frame-wide statistics, never axis by axis. Scale also
changes the failure mode: truncation falls from $41\%$ to $18\%$ at High and to
zero at Low and Medium, which is what makes trace length harmful at low effort
and dominant at high effort in the $120$b block of
Table~\ref{tab:bo64transfer}.

\subsection{RQ2: The Routing Proxy and What It Reads}
\label{app:rq2}

\begin{figure}[t]
\centering
\includegraphics[width=0.35\columnwidth]{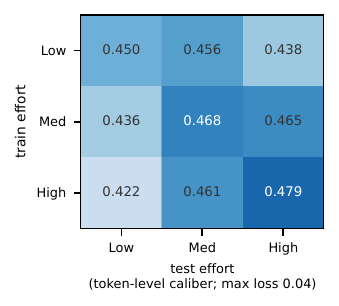}
\caption{Cross-effort transfer of the instantaneous routing$\rightarrow$\jsf{}
map (the token-level version): training the map at one effort setting and testing
at another loses at most $0.04$ reconstruction correlation.}
\label{fig:transfer}
\end{figure}

\begin{figure}[t]
\centering
\includegraphics[width=0.48\columnwidth]{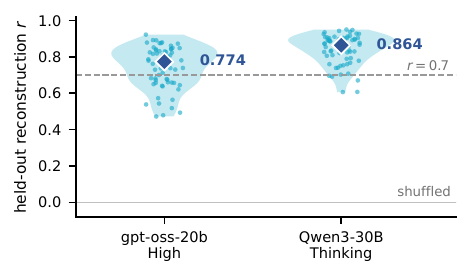}
\caption{Per-axis held-out reconstruction for the two settings whose $64$-axis
vectors are archived: the main figure reports medians and counts for all eight
model--effort cells, this one shows the underlying distributions (dashed line
$r{=}0.7$; the shuffled-routing control sits at zero).}
\label{fig:peraxis}
\end{figure}

\paragraph{Qwen-Instruct reconstruction.}
The archived Qwen reconstruction covers the Thinking variant. For Instruct we
replayed the identical protocol --- three-layer gate-weighted usage spectrum
$\rightarrow$ run-mean \jsf{}, question-held-out four-fold ridge, per-axis Pearson
$r$, row-permutation control --- on that model's own bank and frame, obtaining
median $r=0.833$ with $58/64$ axes above $0.7$ and a permuted control at $-0.019$.
The same code reproduces the archived Thinking values exactly ($0.864$, $59/64$,
permuted $-0.022$), so the two variants are on the same footing in
Figure~\ref{fig:proxy}a.

\paragraph{The proxy inherits the readout's label structure.}
The main text reports R64's outcome-AUC increment over matched token occupancy
under the RQ1 label convention. Under the finished-answer convention of
Appendix~\ref{app:labels} the High-effort increment crosses zero for R64 exactly as
it does for the lens itself, at Low and Medium both remain positive: the proxy
tracks \jsf{} where \jsf{} is informative and where it is not, which is what a
faithful reconstruction should do.

\paragraph{Predicting from raw routing without the bottleneck.}
R64 is a deterministic linear function of the routing spectrum, so an
unconstrained predictor on the raw spectrum can never have a lower attainable
ceiling: by the data-processing inequality the direct arm's supremum accuracy is
at least R64's, with equality only if the reconstruction is a sufficient
statistic. That ordering is structural. How much outcome-relevant routing
information the bottleneck actually discards is then a measurable quantity, and
under the deployment conditions of Table~\ref{tab:bo64transfer} --- frozen
cross-benchmark transfer, one shared regularizer, finite training sets --- it
measures approximately zero: the question-paired gap (direct minus R64) averages
$-0.1$ points over the eight settings, the direct arm leads in three of eight,
every interval except one crosses zero, and the one exception favors R64, $-3.9$
$[-7.5,-0.3]$ on Qwen-Instruct. The mechanism is a ceiling--variance trade: the
direct arm estimates one outcome direction in $99$ or $387$ dimensions, while
R64's first stage fits routing to $64$ dense \jsf{} targets --- an auxiliary task
that explains state-related routing variation before any outcome label is seen
--- and its second stage estimates one direction in $64$; Qwen-Instruct, the cell
with the largest variance pressure, is where the stabilization shows. Routing
through the \jsf{} bottleneck therefore costs nothing in deployment accuracy, and
its coordinates carry names, which the raw spectrum does not: the module--axis
correspondences below and the intervention hypotheses of RQ5 both consume named
axes. The three compressions studied in this paper differ only in the direction
they compress along: supervised by \jsf{}, the bottleneck loses nothing
measurable; chosen by variance (the PCA basis of Appendix~\ref{app:rq1}), it
differs from the frame by estimation efficiency alone; chosen by reconstruction
under nonnegativity (the module selector below), it drops the predictive
direction before any supervised step runs.

\paragraph{What the axes mean.}
Single-token axis names are unreliable, so each axis is characterized by its
\emph{bipolar vocabulary neighbor family}: the token neighborhoods of its
positive and negative poles. Families are semantically coherent across languages
(\emph{stays}$\rightarrow$\emph{remains}/\emph{keeps};
\emph{pursuit}$\rightarrow$\emph{quest}/\emph{commitment};
\emph{caution}$\rightarrow$\emph{careful}/\emph{warning}). That the unit of
meaning is the direction rather than the word is visible in a minimal pair: axes
\#8 (\emph{Intersection}) and \#26 (\emph{intersection}) anchor on the same
surface word yet carry opposite outcome loadings. Screening marks axes that are read as state markers rather than literally:
axis \#44 is a drift-off-the-English-manifold axis (its poles
are non-English script versus English function words); a cluster of digit axes
is an echo hazard; and most axes have garbled negative poles, i.e.\ they are
unidirectional features. Where axes track correctness, the sign structure is
systematic --- keep/advance/continuation families load positive, and
alternatives/meta-planning/complexity families load negative --- a
\emph{resolution-like configuration} (settled goal-pursuit versus unresolved
deliberation) rather than reflection intensity alone.

\paragraph{The eight routing modules.}
Table~\ref{tab:modules} names all eight modules of the token-level factorization
by enrichment. For each module we take the positions in its top activation decile,
compare the token distribution there against the distribution over all positions,
and list the tokens with the highest ratio subject to a minimum count of $40$ in
the high-activation set. The statistic is bounded above by $10$, the inverse of
the decile, so $\times10$ means every occurrence of that token in the sample lies
inside the module's top decile. Ratios are used rather than the most frequent tokens at peak activation, because
raw frequency is dominated by common tokens regardless of the module. The matched axis is the one of the $64$ with the largest absolute
correlation to the module's activation across positions; two modules ($4$ and $6$)
select the same axis with opposite signs, so the modules are distinct in
vocabulary rather than in nearest axis. Modules $0$, $1$ and $7$, in bold, are the
three discussed in RQ2; the first two are the ones traced in
Figure~\ref{fig:modules}.

\begin{table*}[t]
\centering
\footnotesize
\setlength{\tabcolsep}{4pt}
\begin{tabular}{clp{0.31\textwidth}p{0.235\textwidth}lr}
\toprule
Mod. & experts & most enriched tokens ($\times$ ratio) & what it fires on & matched axis & $r$ \\
\midrule
\textbf{0} & 8,23,5,29,16 & \texttt{+b}$\times$7.9, $\theta$$\times$7.8, \texttt{(a}$\times$7.2, \texttt{R}$\times$7.2, \texttt{d}$\times$7.1 & algebraic symbols: variables and expression pieces & \#15 symbol & $+0.48$ \\
\textbf{1} & 11,17,30,13,7 & \texttt{But}$\times$10.0, \texttt{Now}$\times$9.7, \texttt{).}$\times$9.6, \texttt{Wait}$\times$9.6, \texttt{Alternatively}$\times$9.6 & turn markers and paragraph ends & \#44 drift & $+0.59$ \\
2 & 16,31,25,0,8 & \texttt{ings}$\times$9.8, \texttt{letters}$\times$9.8, \texttt{path}$\times$9.8, \texttt{solutions}$\times$9.8, \texttt{blocks}$\times$9.8 & combinatorial objects: discrete structure nouns & \#7 alternation & $-0.45$ \\
3 & 7,29,8,12,18 & \texttt{$>$=}$\times$9.5, \texttt{..}$\times$8.9, \texttt{$>$}$\times$8.9, \texttt{$<$}$\times$8.8, \texttt{)/(}$\times$8.4 & inequalities, fractions and exponents & \#54 numeral & $+0.33$ \\
4 & 13,17,6,20,3 & \texttt{The}$\times$10.0, \texttt{find}$\times$10.0, \texttt{using}$\times$10.0, \texttt{use}$\times$9.8, \texttt{on}$\times$9.6 & method statements (``use \dots{} to find \dots{}'') & \#29 named-object & $+0.45$ \\
5 & 2,17,15,5,1 & \texttt{bis}$\times$10.0, \texttt{equ}$\times$10.0, \texttt{quadr}$\times$9.8, \texttt{.e}$\times$9.8, \texttt{non}$\times$9.6 & geometric word roots (bisect, quadr-, hex-) & \#30 difficulty & $-0.21$ \\
6 & 22,12,2,16,10 & $\approx$$\times$9.3, \texttt{202}$\times$6.9, \texttt{000}$\times$6.4, \texttt{*}$\times$5.3, \texttt{=}$\times$5.1 & numeric approximation and round numbers & \#29 named-object & $-0.30$ \\
\textbf{7} & 0,21,25,17,30 & \texttt{We}$\times$10.0, \texttt{might}$\times$10.0, \texttt{they}$\times$10.0, \texttt{will}$\times$10.0, \texttt{cannot}$\times$10.0 & modal judgement: what is possible or required & \#20 optionality & $+0.38$ \\
\bottomrule
\end{tabular}
\caption{The eight expert modules of the token-level factorization
(gpt-oss-20b, layer $20$, $168$k token positions from $420$ rollouts across the
twelve tier $\times$ benchmark banks; NMF with eight components, run means
removed). Enrichment ratios are capped by the top-decile construction at $10$.
Subword fragments are shown as tokenized. The correlations are from this
$168$k-position run; the main text quotes the independent $347$k-position run,
which agrees with it to within $0.01$ on every module.}
\label{tab:modules}
\end{table*}

\paragraph{Reading the modules, and what that reading assumes.}
At the top of the enrichment scale the ratio saturates: \emph{might}, \emph{will},
\emph{cannot} and \emph{should} reach $\times10$ for Module~7, with \emph{must} at
$\times9.9$. An independent resample ($168$k positions against the $347$k of the main
text) returns the same eight modules, with identical top-five expert sets in seven of
them and one differing in its fifth expert, and reproduces every axis correlation to
within $0.01$.

The axis paired with a module is the one of the $64$ with the largest absolute
correlation, so the pairing is selected by the data even though the axis's name is
not. Finally, axis \#44's word family is non-English (Bengali, transliterated
\emph{t\=ahale}/\emph{\=amr\=a}, ``so''/``we''), so by our naming discipline it is
read as a drift-off-the-English-manifold state marker rather than literally; the
enriched vocabulary of the module that tracks it --- \emph{But}, \emph{Wait},
\emph{Alternatively} --- is what makes that reading specific.

\paragraph{Why the modules are not used as the selector's features.}
The modules of Figure~\ref{fig:modules} are interpretable, so it is natural to ask why
the deployment arms use the $99$-dimensional usage spectrum instead. The reason is that
the factorization optimizes a different objective from the one selection needs.
Nonnegative factorization solves
$\min_{U,Z\ge0}\lVert X-ZU^{\top}\rVert_F^2$, which spends its $K$ components on
directions of large routing variance, whereas a selector needs the direction of
covariance with the outcome $y$. Once $z=U^{\top}x$ is fixed, the supervised stage sees
only the projection of $x$ onto $\mathrm{span}(U)$, so
$R^2 \le \mathrm{corr}^2\!\big(y,\,P_{\mathrm{span}(U)}x\big)$: any predictive
direction lying in $\mathrm{span}(U)^{\perp}$ is discarded before supervision begins,
and a dense ridge has no such bottleneck. Routing makes the bottleneck worse in two
ways. It is compositional, since each token dispatches to four of $32$ experts with
gate weights that nearly sum to one, so the informative quantities are contrasts of the
form ``$A$ rather than $B$''; a nonnegative basis cannot express a contrast in one
component and must form it as a difference of two components that the sum-to-one
constraint makes highly correlated. And nonnegative components are themselves
positively correlated by construction, which leaves the downstream design matrix
collinear and its coefficients high-variance, so a direction fitted on one benchmark
points inconsistently on the next.

The sweep bears this out. Replacing the spectrum by a $K$-module bottleneck and
refitting both stages, the supervised nonnegative bottleneck, which chooses $U$ to
explain \jsf{}, reconstructs better than the unsupervised one at all twelve
$(\text{setting},K)$ pairs ($0.529$--$0.668$ against $0.468$--$0.580$), which is what the
bound predicts once $P_U$ is aimed at the target. On the downstream selection task the
unsupervised bottleneck at $K{=}8$ falls \emph{below} a random pick at Low ($34.2$
against $35.3$) and at Medium ($65.8$ against $66.3$), the signature of an unstable
rather than an absent direction, and recovers only as $K$ grows enough to readmit the
predictive direction ($35.0$ and $69.4$ at $K{=}16$). We therefore read the modules as
an account of how routing is organized, and keep the full spectrum as the feature set.

\paragraph{Routing specializes in process state.}
Routing is substantially more informative about trajectory quality and
completion dynamics than about exact answer content. Predicting \emph{which answer string} a trace will
commit to from routing alone performs at chance ($0.493$, and coarse top-$k$
expert selections show no stable answer-commitment fingerprint), whereas
predicting \emph{whether a trajectory will be correct} --- the task used
throughout RQ2--RQ4 --- is where routing matches the lens. This specialization
matches routing's downstream role as a process monitor for confidence
estimation and compute allocation.

\subsection{RQ3: Branch Selection and Vote Weighting}
\label{app:rq3}

\begin{table*}[!t]
\centering
\small
\setlength{\tabcolsep}{5pt}
\begin{tabular}{lccccc c}
\toprule
 & \multicolumn{3}{c}{gpt-oss-20b} & \multicolumn{2}{c}{Qwen3-30B-A3B} & \textbf{pooled} \\
\cmidrule(lr){2-4}\cmidrule(lr){5-6}
 & Low & Med & High & Think & Inst & \\
questions & 198 & 198 & 198 & 198 & 198 & 990 \\
\midrule
random pick & 56.3 & 64.4 & 59.3 & 71.1 & 66.3 & 63.5 \\
J64 & \textbf{59.6}\gp{3.3} & \textbf{71.7}\gp{7.4}\sig{} & 67.2\gp{7.9}\sig{} & \textbf{72.7}\gp{1.7} & 69.2\gp{2.9} & \textbf{68.1}\gp{4.6}\sig{} \\
R64 (routing) & \textbf{59.6}\gp{3.3} & 69.7\gp{5.3}\sig{} & 65.2\gp{5.8}\sig{} & 70.7\gm{0.4} & \textbf{73.7}\gp{7.4}\sig{} & 67.8\gp{4.3}\sig{} \\
token occupancy & 55.1\gm{1.2} & 66.7\gp{2.3} & \textbf{67.7}\gp{8.4}\sig{} & 72.2\gp{1.2} & 64.6\gm{1.7} & 65.3\gp{1.8} \\
length & 50.5\gm{5.8}\sig{} & 64.1\gm{0.2} & \textbf{67.7}\gp{8.4}\sig{} & 70.7\gm{0.4} & 63.6\gm{2.7} & 63.3\gm{0.1} \\
DeepConf & 54.5\gm{1.7} & 63.1\gm{1.2} & 60.6\gp{1.3} & 70.2\gm{0.9} & 64.1\gm{2.2} & 62.5\gm{0.9} \\
\midrule
\emph{(oracle)} & 97.0 & 97.0 & 89.9 & 91.4 & 92.9 & 93.6 \\
\bottomrule
\end{tabular}
\caption{Single-branch selection on GPQA (\%), best-of-$64$: a second task domain,
$198$ graduate-level science questions $\times$ $64$ rollouts per setting,
question-held-out folds and question-residualized features
(Appendix~\ref{app:rq3}). Parenthesized values are the gain over the random-pick
reference of the same column; \textbf{pooled} is the single rate over all $990$
question--setting units. Bold $=$ best per column.}
\label{tab:gpqasel}
\end{table*}

\paragraph{GPQA: protocol and boundaries.}
GPQA is a single benchmark, so the cross-benchmark frozen-transfer protocol of
Table~\ref{tab:bo64transfer} does not apply; every learned arm in
Table~\ref{tab:gpqasel} is instead fitted with question-held-out folds within the
setting, under the same regularizer, on \emph{question-residualized} features, each
feature minus its own question's pool mean. Residualization is what makes the
comparison meaningful on science questions: token occupancy otherwise encodes which
question this is, and a selector scores by learning that hard questions fail more
often; what remains after residualizing is how a rollout differs from its siblings,
which is what selection actually needs. oss-20b
High is the only column where the behavioral arms clear zero as well, with trace
length among its strongest arms, the signature of the unfinished-rollout failure mode
established for the mathematics benchmarks. In the
remaining columns only the internal arms ever clear zero, \jsf{} and R64 at Med and
R64 at Inst, and at Low and Think no arm clears zero in the positive direction. The
paired contrast between \jsf{} and token occupancy crosses zero in every individual
setting, so on this domain the readout's advantage is that it is the only channel
positive throughout rather than a margin in any single column.

\paragraph{A selector tracks the dominant failure mode.}
Where a behavioral baseline beats the internal arms in
Table~\ref{tab:bo64transfer}, the reversal is legible from the failure mode of the
setting. At $120$b the Low and Medium settings have \emph{zero} truncation, so failures
are pure capability errors and trace length is actively harmful, at $-13.8$ and $-8.3$
points. At High, truncation returns at $18\%$ and length becomes the strongest
single-trace arm. The reversal that separates model families in the Qwen block of the
main table therefore reappears inside a single model across its effort dial, which is
why we read length as a mode-specific baseline rather than a competitor.

\paragraph{Consensus on the rescue slice, and the recoverable ceiling.}
\emph{On the oracle-conditioned rescue slice the consensus family stops helping.} The
weighted-vote variants do not improve over random single-branch selection on average
there --- they straddle zero, and DeepConf-weighted voting falls $4.3$ points below it
--- so the internal score contributes something consensus does not, rather than a
better way of counting votes; trace length is actively harmful on the same slice
($5.4\%$, $-3.4$\sig{}), because on failed questions the extreme-length branches are
mostly degenerate. \emph{The rescue headroom is real rather than saturated.} Only
$45$--$75\%$ of those pools contain any correct branch at all, and against that
ceiling \jsf{} converts about a quarter of what is recoverable.

\paragraph{Pool composition and the unconditional protocol.}
All selection numbers in the paper are unconditional --- every question of every
test set enters, including pools that no selector can lose and pools that none
can win. Out of $30$ questions per benchmark, the counts of
all-correct\,/\,all-wrong\,/\,mixed pools at $N{=}64$ are $0$--$1$\,/\,$4$--$9$%
\,/\,$21$--$26$ at Low effort, $2$--$7$\,/\,$0$--$2$\,/\,$23$--$26$ at Medium,
and $3$--$10$\,/\,$1$--$3$\,/\,$18$--$25$ at High. The two protocols therefore
differ in opposite directions by effort: at Low the all-wrong pools pull the
unconditional average \emph{down} relative to a mixed-pool recomputation (e.g.\
HMMT-25 at $N{=}16$: \jsf{} $0.245$ unconditional vs $0.350$ on mixed pools),
while at High the all-correct pools pull it \emph{up} (BRUMO-25: $0.813$ vs
$0.757$). Recomputing every arm on mixed pools only leaves the arm ordering unchanged at
all three efforts and makes DeepConf's high-effort collapse deeper ($0.44$--$0.61$
at $N{=}64$). What the composition does bound is the headroom: an oracle that always
picks a correct branch when one exists reaches only $0.56$--$0.76$ at Low
($N{=}16$), and at High \jsf{} already captures about six-tenths of the
available oracle-minus-random gap.

\paragraph{Rescue by setting.}
Table~\ref{tab:rescuefull} breaks the main-text rescue result --- \jsf{}
$17.0\%$ versus $8.7\%$ for a random pick over the $223$ vote-failure questions
--- down by individual setting.

\begin{table*}[!t]
\centering
\scriptsize
\setlength{\tabcolsep}{1.6pt}
\resizebox{\textwidth}{!}{%
\begin{tabular}{l cccc cccc cc c}
\toprule
 & \multicolumn{4}{c}{gpt-oss-20b} & \multicolumn{4}{c}{gpt-oss-120b} & \multicolumn{2}{c}{Qwen3-30B-A3B} & all \\
\cmidrule(lr){2-5}\cmidrule(lr){6-9}\cmidrule(lr){10-11}\cmidrule(lr){12-12}
 & Low & Med & High & \emph{avg} & Low & Med & High & \emph{avg} & Think & Inst & \textbf{pooled} \\
failed qs & 64 & 20 & 18 & 102 & 50 & 19 & 6 & 75 & 11 & 35 & 223 \\
\midrule
random pick & 9.0 & 12.3 & 4.2 & 8.5 & 8.4 & 16.8 & 6.8 & 10.6 & 4.1 & 6.5 & 8.7 \\
J64 & \textbf{14.1}\gp{5.1}\sig{} & \textbf{23.3}\gp{11.0} & \textbf{5.6}\gp{1.4} & \textbf{14.3}\gp{5.8} & \textbf{17.3}\gp{9.0}\sig{} & 24.6\gp{7.8} & 27.8\gp{21.0} & \textbf{23.2}\gp{12.6} & \textbf{21.2}\gp{17.1}\sig{} & 17.1\gp{10.6}\sig{} & \textbf{17.0}\gp{8.3}\sig{} \\
R64 & 9.9\gp{0.9} & 23.3\gp{11.0} & 3.7\gm{0.5} & 12.3\gp{3.8} & 11.3\gp{3.0} & 26.3\gp{9.5} & 22.2\gp{15.5} & 20.0\gp{9.3} & 15.2\gp{11.0} & 17.1\gp{10.6}\sig{} & 14.1\gp{5.3}\sig{} \\
token occupancy & 8.9\gm{0.2} & 10.0\gm{2.3} & 5.6\gp{1.4} & 8.1\gm{0.4} & 13.3\gp{5.0} & 19.3\gp{2.5} & 11.1\gp{4.3} & 14.6\gp{3.9} & 12.1\gp{8.0} & \textbf{18.1}\gp{11.6}\sig{} & 12.3\gp{3.5}\sig{} \\
length & 0.0\gm{9.0}\sig{} & 10.0\gm{2.3} & 0.0\gm{4.2}\sig{} & 3.3\gm{5.2} & 0.0\gm{8.4}\sig{} & 10.5\gm{6.2} & \textbf{33.3}\gp{26.6} & 14.6\gp{4.0} & 9.1\gp{5.0} & 14.3\gp{7.8} & 5.4\gm{3.4}\sig{} \\
DeepConf & 12.5\gp{3.5} & 10.0\gm{2.3} & 5.6\gp{1.4} & 9.4\gp{0.8} & 8.0\gm{0.4} & \textbf{36.8}\gp{20.1}\sig{} & 16.7\gp{9.9} & 20.5\gp{9.9} & 9.1\gp{5.0} & 14.3\gp{7.8} & 13.0\gp{4.3}\sig{} \\
\bottomrule
\end{tabular}%
}
\caption{Rescue on the questions where plain majority voting scores zero, by
setting (\%). The protocol is identical to Table~\ref{tab:bo64transfer} ---
frozen cross-benchmark transfer, one uniform regularizer, deterministic single
pick from the full pool of $64$, averaged over the same $12$ transfer cells per
setting --- restricted to the failed questions of each test set.
Parenthesized values are question-paired gains over a random pick in the same
column, a star marks a clustered CI excluding zero, and \emph{avg} columns are
unweighted means over a model's three effort settings. Individual settings hold
only $6$--$64$ failed questions, so single cells indicate direction; pooled over
all $223$ failed questions \jsf{} rescues $17.0\%$ against $8.7\%$ for a random
pick ($+8.3$\sig{}$[+5.3,+11.3]$) and R64 reaches $14.1\%$
($+5.3$\sig{}$[+2.3,+8.5]$), while length falls to $5.4\%$
($-3.4$\sig{}$[-6.4,-0.1]$).}
\label{tab:rescuefull}
\end{table*}

\subsection{RQ4: Online Compute Allocation}
\label{app:rq4}

\paragraph{The two readout forms against each other.}
Table~\ref{tab:cusum} compares both readout forms against the sibling-permuted
control, which is the comparison the online claim rests on. Against the lens
directly, R64 trails on gpt-oss-20b High by $+1.4$\sig{} to $+2.7$\sig{} points,
on Qwen the two are equivalent, and at the cheapest target the ordering inverts,
with R64 leading \jsf{} by $0.9$\sig{} at $1.2\times$. Reading hidden states
therefore buys a margin only at the higher budgets of one model group.

\subsection{RQ5: Mechanism-Matched Routing Interventions}
\label{app:rq5}

\begin{table*}[t]
\centering
\footnotesize
\setlength{\tabcolsep}{3.5pt}
\begin{tabular}{%
>{\raggedright\arraybackslash}p{0.150\textwidth}%
>{\raggedright\arraybackslash}p{0.105\textwidth}%
>{\raggedright\arraybackslash}p{0.180\textwidth}%
>{\raggedright\arraybackslash}p{0.140\textwidth}%
>{\raggedright\arraybackslash}p{0.110\textwidth}%
>{\raggedright\arraybackslash}p{0.180\textwidth}}
\toprule
Diagnosed process & \jsf{} axes & Routing component & Intervention experts &
Predicted change & Observed change \\
\midrule
Excessive case splitting, failure to commit &
case-splitting bundle \#20/\#7/\#16 &
experts ranked by write-vector \emph{advantage} on the bundle direction, layers
$8$--$19$ &
top two per layer, state-dependent &
more splitting, less commitment &
``Case $n$'' enumeration ($100$ headers in one run; $0$ of $160$ sham runs),
non-termination (cap $0.98$), accuracy $\rightarrow 0.000$ \\
\addlinespace
Hovering and probing instead of operating &
hover \#22, requirement \#1, probe \#8 \emph{vs} operation \#47 &
expert group whose gated usage separates incorrect from correct trajectories
(within-question, length-controlled) &
L12-E23, L12-E12, L12-E24, L16-E11, L20-E0; suppressed; $1.2\%$ of gated mass &
less probing, faster exact execution &
shorter generations ($-45.9$\sig{}; sham $+47.6$\sig{}), symbolic derivation
replaces numerical guessing, answer becomes correct \\
\bottomrule
\end{tabular}
\caption{The two mechanism-matched interventions of RQ5, from the diagnosed process
to the observed change.}
\label{tab:rq5link}
\end{table*}

\paragraph{Target selection, its relation to R64, and scope.}
Table~\ref{tab:rq5link} lays the two interventions out end to end. The cases enter the
routing substrate from opposite sides: Case~1 selects experts \emph{geometrically}
from a \jsf{} direction: at each layer $l\in[8,19]$ experts are ranked by the
advantage of their write vector along the bundle direction,
$a_e=\big(w_e(h)-\sum_j p_j w_j(h)\big)\cdot\hat{w}_{\text{bundle}}$ with the
second term the currently gated mixture, and the top two per layer are edited,
state-dependently. Case~2 selects them
\emph{statistically} from routing usage --- L12-E23, L12-E12, L12-E24, L16-E11 and
L20-E0, together $1.2\%$ of the gated mass across the three recorded layers --- and
then reads their \jsf{} profile. The Case~2 group also coincides with the unsupervised structure of RQ2. Factorizing
per-token gate usage into eight modules separately at each recorded layer, with no
correctness label anywhere in the procedure, places all three layer-$12$ targets in one
module, experts $\{7,9,12,22,23,24,26,28,29\}$, two of them with all of their loading
on it and the third with $43\%$; three experts drawn at random land in a common module
with probability $0.040$, or $0.027$ if drawn distinct. The layer-$16$ target carries
$100\%$ of its loading on its module and the layer-$20$ target $88\%$. The causally
effective group is therefore a coalition the routing substrate already contains, not a
set assembled by the outcome statistics that selected it.

Neither selection rule is a ranking of R64's reconstruction
coefficients, and the three readings measure different quantities: fitting the
same ridge and ranking the $96$ expert features by their weight on an axis puts the
experts we intervene on well down the list. On the hover axis the expert whose write
vector loads on it in five of five reference sets ranks $80$th of $96$ by
reconstruction weight, and none of that axis's five largest reconstruction
contributors is an intervention target; on Case~1's three axes the targets rank
between $14$th and $93$rd. The three questions differ: reconstruction weight asks which
experts \emph{predict} the readout and is dominated by high-usage experts that carry
variance, module membership asks which experts are \emph{used together}, and the
advantage rule asks which experts \emph{move} the readout. A target can rank $80$th of
$96$ on the first and be a core member on the second without contradiction.

In the Case~1 push (ten AIME-24 problems $\times$ $16$ rollouts per arm, paired by
question and seed) the unmodified model does produce occasional ``Case $n$''
headers --- one to five in $7$ of $160$ runs --- so the specificity claim is about
the targeted arm against the $160$ equal-dose sham runs, in which no header occurs
at all. In Case~2 the profile is stable per expert across five
independent reference sets, but only the layer-$20$ member has the
hover/probe/requirement reading quoted in the main text; the other four separate the
outcome classes statistically without that profile, and the probing direction (\#8)
is an axis whose lens label grounding rejected, so we read it as a state marker
rather than literally. The dose window is narrow --- expert-level constants of
$0.3$--$0.8$ work and larger doses return to baseline --- and the arena throughout is
gpt-oss-20b at Low effort on competition mathematics.

\paragraph{Arm-level accuracies.}
Table~\ref{tab:rq5acc} gives every arm on the two samples we ran. On twelve problems
($12\times16$ rollouts $\times$ seven arms, $1{,}344$ trajectories) promoting the
outcome-positive experts at dose $0.3$ reaches $0.411$ and suppressing the negative
group at $0.8$ reaches $0.406$, against $0.380$ for no-op and $0.359$ for the equal-dose
sham; every treatment arm is at or above no-op and the sham is the only arm below it;
among the single-direction treatment arms the treatment--sham separation is $+3.6$
to $+5.2$ points. On the three case-study
problems suppression reaches $0.625$ against $0.542$ for no-op and $0.500$ for the sham,
a $+12.5$-point separation. The
per-problem breakdown shows where the gain comes from: on the rescued problem the
sham also recovers $2$ of $16$, and what separates the arms is that suppression leaves
the problems the model already solves intact or improves them ($12/16$ and $14/16$
becoming $12/16$ and $16/16$) while the sham degrades both ($10/16$, $12/16$).

\begin{table}[t]
\centering
\small
\setlength{\tabcolsep}{4pt}
\begin{tabular}{lcc}
\toprule
Arm & twelve & three \\
 & problems & case studies \\
\midrule
no-op & $0.380$ & $0.542$ \\
promote pos.\ ($0.3$) & \textbf{0.411}\gp{3.1} & --- \\
suppress neg.\ ($0.8$) & $0.406$\gp{2.6} & \textbf{0.625}\gp{8.3} \\
suppress neg.\ ($0.3$) & $0.396$\gp{1.6} & --- \\
promote pos.\ ($0.8$) & $0.396$\gp{1.6} & $0.604$\gp{6.2} \\
both ($0.8$) & $0.380$\gz{0.0} & $0.583$\gp{4.2} \\
sham ($0.8$, random) & $0.359$\gm{2.1} & $0.500$\gm{4.2} \\
\bottomrule
\end{tabular}
\caption{Accuracy of every intervention arm on the two samples (gpt-oss-20b, Low
effort, $16$ rollouts per problem per arm; $1{,}344$ and $240$ trajectories).
Parenthesized values are points over no-op, \textcolor{gain}{green} above it and
\textcolor{loss}{red} below. Every treatment arm evaluated on a given sample is at or above no-op; the sham is
the only arm below it. A constant is subtracted from, or added to, the router
logits of the selected experts at every token; on the three case studies
accuracies are multiples of $1/48$, so the arm-level intervals there are too
coarse to separate the arms and only the direction is read. A trigger-gated
variant that suppresses only at tokens where a targeted expert is naturally
selected reproduces the twelve-problem result ($0.401$ against $0.406$).}
\label{tab:rq5acc}
\end{table}

\paragraph{Expert-level target selection and the intervention arms.}
The targets of the second study are chosen before any intervention, from usage
statistics alone. On $5{,}834$ finished Low-effort runs we standardize each
expert's usage rate within question and compare correct against incorrect
rollouts. The five most negative are L12-E23 ($d=-0.285$\sig{}), L12-E12
($-0.255$\sig{}), L12-E24 ($-0.248$\sig{}), L16-E11 ($-0.233$\sig{}) and L20-E0
($-0.229$\sig{}); the three most positive are L12-E19 ($+0.192$\sig{}), L16-E29
($+0.149$\sig{}) and L12-E14 ($+0.147$\sig{}). Controlling for trace length leaves
the five negative effects essentially unchanged
($-0.290$/$-0.250$/$-0.240$/$-0.227$/$-0.229$), and the two strongest positive
effects likewise ($+0.185$/$+0.148$; the control was not recomputed for L12-E14),
so these are not length artifacts. The suppression arms act on the five negative
experts; the promotion arms act on the four experts L12-E19, L16-E29, L12-E14 and
L16-E19, the fourth fixed in the arm at construction time with an effect size
outside the archived top-three ranking. Splitting each run into
fifths, correct rollouts use the negative group at
$0.070$/$0.023$/$0.021$/$0.039$/$0.093$ and incorrect ones at
$0.116$/$0.085$/$0.085$/$0.097$/$0.138$: the gap is concentrated in the middle of the
run, the core computation. The write-vector profiles are complementary --- L20-E0
loads on \emph{keep hovering} and \emph{probe a direction} and against
\emph{intersection}, \emph{intersect verb} and \emph{case split}, while the
positive L16-E29 loads on \emph{intersection} ($+14{,}750$) --- which is the basis
for reading the intervention as ``stop hovering, do the operation''. The
arm-by-arm outcome on the three case-study problems is the second column of
Table~\ref{tab:rq5acc}.

\paragraph{Rollout transcripts and effect intervals.}
In the case-amplification study, on AIME-24 P6 the targeted arm proceeds through
``\emph{Case 72} $\dots$ \emph{Case 86}'' while the equal-dose sham repeats a
single sentence with no case structure: both arms lose the same accuracy, and only
the targeted one loses it in the shape the diagnosis names. In the suppression
study the length change is $-45.9$ tokens $[-75.7,-17.3]$\sig{} under treatment
against $+47.6$ $[+13.5,+82.9]$\sig{} under the equal-dose sham. On AIME-24 P5,
the inradius of a tetrahedron with equal opposite edges, the unmodified trajectory
converts to decimals early and then guesses at a radical, ``\emph{$\dots$
$r=1.162$ $\dots$ Suppose exact $r=\sqrt5/?$ $\dots$ Not nice $\dots$ Given time,
guess answer 12?}'', whereas with the diagnosed experts suppressed the same
problem is carried in exact symbolic form throughout, ``\emph{an isosceles
tetrahedron with opposite edges equal $\dots$ $V=160/3$ $\dots$
$r=20\sqrt{21}/63$ $\dots$ $m+n+p=104$}''.

\section{Reproducibility: Constructions and Protocols}
\label{app:repro}

This section specifies every construction the main text uses. Values are the ones
in the code that produced the reported artifacts.

\subsection{The \jsf{} Frame and the Two Needles}
\label{app:frame}

\begin{table*}[t]
\centering
\small
\begin{minipage}{0.94\textwidth}
\hrule\vspace{2pt}
\textbf{Algorithm 1} \quad \jsf{} frame construction (once per model; no outcome,
effort or difficulty labels)\vspace{2pt}
\hrule\vspace{3pt}
\begin{enumerate}\itemsep1pt \parskip0pt \topsep0pt
\item \textbf{Lens.} Fit a per-source-layer Jacobian lens
\citep{gurnee2026verbalizable}: for source layer $\ell$,
a linear map $J_\ell$ carrying the residual stream at $\ell$ into the final-layer
basis, estimated by regression with one-hot cotangent injection at every token
position from index $16$ onward. Read the frame at one layer $L$ per model:
$L{=}20$ of $24$ for gpt-oss-20b, $L{=}23$ of $36$ for gpt-oss-120b, $L{=}34$ of
$48$ for Qwen3-30B-A3B (both variants). This is a fitted lens, not a bare
unembedding logit lens; for gpt-oss-20b we use the published checkpoint of
\citet{gurnee2026verbalizable} hosted on Neuronpedia \citep{neuronpedia}, and the
gpt-oss-120b and Qwen lenses are fitted with the same estimator.
\item \textbf{Dictionary.} $D \leftarrow$ rows of $W_U J_L$, each row $\ell_2$%
-normalized; rows of special tokens are zeroed. Every non-special vocabulary item
is a candidate direction; there is no hand-built word list.
\item \textbf{State pool.} Replay the first $6$ problems $\times$ first $2$
rollouts of each of the four benchmarks and collect hidden states at $L$ from token
$16$ onward; sample $400$ states with seed $0$.
\item \textbf{Atom seeding} (one of two simple statistics; every other step is
identical). \emph{(a) Sparse-coding mass:} code each sampled state over $D$ with
non-negative orthogonal matching pursuit, $k{=}25$ atoms per state, accumulate each
atom's non-negative coefficient mass over the $400$ states, and keep the
$512={64}\times8$ atoms of highest mass. \emph{(b) Lens-decode frequency:} count
each token's appearances among the top-$20$ lens decodings of the sampled states and
keep the $400$ most frequent. The seeder is fixed per model at construction time from the construction-pool
diagnostics described below, which use no outcome, effort or difficulty label:
gpt-oss-20b uses (a); gpt-oss-120b and Qwen3-30B-A3B use (b).
\item \textbf{Grouping.} Walk the candidates in decreasing seeder score and greedily open a
new group whenever an atom's cosine to every existing group representative is
$\le0.7$, otherwise add it to the group it exceeds $0.7$ with; keep the first $64$
groups.
\item \textbf{Axes.} Each axis is the seeder-score-weighted mean of its members' dictionary
rows, $\ell_2$-normalized, giving the frame
$A=[a_1\,\cdots\,a_{64}]\in\mathbb{R}^{d\times64}$ whose columns are the axes.
\item \textbf{Readings.} $\jsf{}(h) = A^{+}(h-\mu)$ with $A^{+}$ the
Moore--Penrose pseudo-inverse and $\mu$ the mean state over a fixed four-sentence
neutral corpus.
Readings are raw projection coefficients, and each axis points toward its own token
family because both the sparse code and the group weights are non-negative, so a
reading may take either sign. The ``units'' quoted in RQ5 are these coefficients:
the case-splitting bundle's reading rises by $66.3$ units from Low to High effort,
which is $1.08$ standard deviations of that reading ($sd=61.3$) --- the conversion
between the raw units of RQ5 and the standardized readings plotted elsewhere.
\end{enumerate}
\vspace{2pt}\hrule\vspace{2pt}
\emph{No step uses an outcome, difficulty or effort label, and no axis is manually
selected or edited: the $64$ axes are whatever the chosen seeding statistic and the
$0.7$ threshold produce. The state pool, the grouping procedure and its threshold are
shared by all three models; the seeder and its candidate-bank size ($512$ atoms under
sparse coding, $400$ under decode frequency) are model-specific. Axis \emph{names}
and responsibility tiers are assigned afterwards and change no number. Each frame is
built independently in its own model's space, so axes are not aligned index-by-index
across models.}
\end{minipage}
\end{table*}

\paragraph{Per-model atom seeding.}
The seeder is chosen per model by two construction-pool diagnostics, applied in
order and before any downstream analysis. First, the seeder must be
non-degenerate: its atoms must reconstruct held-out states and cluster into
multi-token families. Second, among non-degenerate seeders, the $64$ axes should
vary \emph{within} a question across sibling rollouts rather than between
questions, a comparison that uses question identity and nothing else. Neither
diagnostic consults an outcome, effort or difficulty label. On gpt-oss-20b the
sparse-coding seeder passes: its axes place $0.173$ of their variance within questions
against $0.164$ for the frequency seeder, and only $5$ of $64$ axes are question-locked
against $9$. On gpt-oss-120b the same seeder drifts onto question topic, with the
within-question share at $0.132$ and $15$ of $64$ axes locked onto axes reading
\emph{Permutation}, \emph{Quaternion}, \emph{gcd}; the frequency seeder is chosen
there. On Qwen3-30B-A3B the sparse coder degenerates in a different way, onto
off-manifold code fragments: a state's $25$ atoms reconstruct only $0.036$ of its
energy, the lowest of the three models, and the median group size after clustering is
$1$, whereas the frequency seeder yields coherent process families
(\emph{maybe/perhaps}, \emph{calculation}, \emph{measurement}) and reconstructs held-out
states at $0.049$ against $0.035$. The choice only moves the seed: on gpt-oss-120b,
where both frames were carried through the full selection and rescue protocols, every
arm keeps its sign and ordering under either seeder, with the single-branch \jsf{} arm
$2.7$ points $[+0.3,+5.1]$\sig{} stronger under the chosen frame ($+7.1$
$[+1.3,+12.9]$\sig{} on the rescue slice) and the R64 and vote-weighted arms
essentially unchanged. Signal-bearing axis counts ($28$ against $22$ at $120$b) are
reported for context only and did not enter the selection.

\paragraph{Frame overlap and what is supervised.}
Frame construction consumes $24$ of the $120$ evaluation questions. All headline
evaluations use all $120$ questions; construction-disjoint controls on the remaining
$96$ are additionally reported for the RQ1 diagnostics (Appendix~\ref{app:rq1}) and
for the RQ3 transfer grid (below), because the frame is a data-dependent
representation even though it is outcome-label-free. Everything built on top of the
frame \emph{is} supervised and is trained only on source questions: the RQ3
selectors and vote weights (correctness labels), the RQ4 prefix score (failure
labels), and the R64 reconstruction map (\jsf{} targets, no outcome labels).

\paragraph{The two coordinates.}
Let $Z$ be the readings column-standardized over the analysis sample ($480$ runs,
stratified over three effort settings $\times$ four benchmarks, seed $0$), and write
$\overline{Z}_S$ for the mean of $Z$ over an index set $S$. With
$B=\{20,7,16\}$ (the case-splitting bundle) and $A=\{45,52,50,0,26,35\}$ (the
arithmetic core),
\[
\text{posture} = \overline{Z}_B-\overline{Z}_A,
\qquad
\text{strain} = Z_{30}-Z_{1},
\]
where axis $30$ is the problem-perception axis and axis $1$ the
constraint-requirement axis. The axis sets were chosen on a random half of the
problems (seed $1$); the $60$ problems of the other half are the held-out
confirmation quoted in RQ1.

\subsection{Labels, the Matched Text Channel and the RQ1 Estimator}
\label{app:labels}

\paragraph{Labels.}
Two outcome conventions appear in the paper, one per use. The outcome label of RQ1
and RQ4 is $1$ if the extracted answer matches the gold answer \emph{and} the run
terminated on its own: an attempt that hits the generation limit delivers nothing, so
the label is ``a correct answer was delivered''. Selection over completed pools (RQ3)
scores a branch by its extracted answer alone, because an answer recovered from a
truncated trace is still a usable pick. On gpt-oss-20b High the two conventions read
the same pool as follows: $59.0\%$ of rollouts finish and score $96.2\%$ when they
do, and truncated rollouts still carry an extractable correct answer $34.3\%$ of the
time, which gives the random pick of Table~\ref{tab:bo64transfer} its $70.9$ and one
uninterrupted rollout in Table~\ref{tab:cusum} its $56.8$. Problem difficulty is $1$ minus the solve rate over that problem's
rollouts at the same effort setting. That estimate uses the same rollouts as the
strain--difficulty correlation, so the correlation is not cross-fitted and shares
sampling noise with its target; this is why we report the held-out-problem
replication ($+0.28$/$+0.35$/$+0.28$) alongside the pooled value.

\paragraph{The matched text channel.}
Within each training fold we count tokens over that fold's rollouts only and keep
the $3{,}000$ most frequent, so the vocabulary never sees held-out questions; it is
rebuilt per model and per effort setting. Out-of-vocabulary tokens are dropped;
punctuation and subword pieces are kept exactly as tokenized. Counts are
$\log(1{+}x)$ transformed and each rollout's vector is $\ell_2$-normalized, after
which the classifier standardizes features using training-fold statistics.

\paragraph{Estimator.}
Logistic regression (lbfgs, $2{,}000$ iterations, no class weighting, no post-hoc
calibration) with inverse regularization $C\in\{1/3,1/30\}$ chosen by mean AUC and
then held fixed for both channels; four question-level folds; features standardized
per fold. AUC is computed as a within-problem paired comparison pooled over
problems, and intervals are problem-level cluster bootstraps ($2{,}000$--$4{,}000$
resamples).

\subsection{R64: Routing Features and the Reconstruction Map}
\label{app:r64}

\paragraph{How each channel is captured.}
Routing is recorded by the generation stack itself: expert assignments and gate
weights are written out as the trajectory is sampled, so the routing features cost
nothing beyond generation. Activations are not exposed there, so \jsf{} readings are
obtained by replaying each finished trajectory through the model under teacher
forcing, with a hook on the residual stream at layer $L$; this is a second forward
pass over every token of the trajectory. That replay also re-records the router
outputs, which is how the two channels are aligned token by token, but the routing
used in the paper is the generation-time record. In-process instrumentation costs,
measured on gpt-oss-20b over $256$ decoded tokens, median of three runs: bare
decode $38.37$ ms per token, $38.44$ with the gate-capture hook, $38.47$ with the
\jsf{}-projection hook, and $38.42$ with both, so each hook sits within run-to-run
noise. One NN-OMP decomposition over the $201$k-atom dictionary, used only when
seeding the frame, costs $43.3$ ms per state.

\paragraph{R64 features and map.}
The usage spectrum reads three MoE layers ($12,16,20$ for gpt-oss-20b), all
top-$4$ assignments kept: per expert, the summed gate weight divided by the number
of tokens in the run ($32$ experts per layer), plus one gate entropy per layer,
computed per token and averaged over the run --- $3\times32+3=99$ features, and
$3\times128+3=387$ for gpt-oss-120b and Qwen3-30B-A3B. The map is fitted differently for its two uses:
RQ2 evaluates a model--effort-specific map with question-held-out folds over that
setting's whole reconstruction corpus (Figure~\ref{fig:proxy}a, eight such cells),
whereas for RQ3 both the reconstruction map and the selector on top of it are fitted
on the \emph{source benchmark alone} and frozen together before transfer: in the code the ridge is
solved on the source benchmark's routing and \jsf{} matrices and then applied
unchanged to each target, so no target rollout contributes to either stage. The
cross-effort experiment shows
a single token-level map transferring between settings at a cost of at most $0.04$,
which suggests but does not establish that one map per model would suffice. The
window-level map of RQ4 shares no parameters with the trajectory-level map. The map
is a single multi-output ridge solved in closed form with a fixed
$\lambda{=}50$ (the ``one uniform regularizer''), inputs standardized, targets
raw, over four question-grouped folds. Per-axis Pearson $r$ is computed on pooled
out-of-fold predictions, and axes with negative $r$ are kept in both the median and
the count above $0.7$. The permutation control permutes routing rows across all
runs.

\subsection{RQ3: Selection, Transfer and Voting Protocols}
\label{app:rq3proto}

\paragraph{Selection, transfer and vote weighting.}
Pools are $64$ rollouts per question. A selector is a logistic regression
($500$ iterations) fitted on one source benchmark and frozen, with the inverse
regularization shared by every learned arm and every setting at $C{=}3$, the one
hyperparameter of Table~\ref{tab:bo64transfer} not selected per source. It was fixed
once, by four-fold question-level cross-validation inside a single benchmark, and
applied unchanged to every arm, setting and target. Selecting inside AIME-24 or
AIME-25 at High effort returns $C{=}3$ under all three criteria we tried (mean
held-out best-of-$64$ accuracy over the learned arms, over the two internal arms, or
for \jsf{} alone); selecting inside HMMT-25 or BRUMO-25 returns values between
$10^{-3}$ and $10^{-1}$, and the selecting benchmark's questions also appear as
transfer targets elsewhere in the grid. Across the global grid $C\in[10^{-3},3]$,
\jsf{}'s effort-averaged best-of-$64$ at Low moves between $35.0$ and $41.9$.
A stricter protocol removes the shared value entirely: selecting $C$ per (setting,
source, arm) by four-fold question-level cross-validation inside the source benchmark
alone, then refitting and freezing, changes the five effort-setting averages by at
most $2.5$ points (\jsf{} $41.7/68.6/75.0/83.6/66.1$ against $41.9/68.1/77.5/85.3/%
65.6$ at $C{=}3$; R64 $34.7/67.5/75.8/87.5/65.3$ against $35.8/68.3/76.1/87.2/64.7$)
and moves the vote-weighted arms by at most half a point. \jsf{} stays above the
random reference in all five settings; the one sign change anywhere is R64 at Low,
$+0.8$ to $-0.4$. The main table keeps $C{=}3$. For R64, the
reconstruction ridge is frozen with the selector. Single-branch selection takes the argmax of
the score over the pool. Vote weighting sums the frozen selector's probability over
branches that agree on an answer and takes the argmax; only terminated branches
with a parseable answer take part. Each cell is the macro-average over the $12$
source$\rightarrow$target pairs ($4$ sources $\times$ $3$ held-out targets), so a
test question is evaluated once per source head and the reported number is not a
single prediction per question; intervals are problem-level bootstraps computed
within each target set and then averaged, which does not model the correlation
induced by re-using a test question across the three source heads --- they should
be read as within-target intervals.

\paragraph{Construction-disjoint check.}
We ran the control. Frame construction consumes problems $0$--$5$ of each
benchmark, so we repeated the five gpt-oss-20b and Qwen transfer settings of
Table~\ref{tab:bo64transfer} with those $24$ questions removed from both selector
fitting and evaluation, leaving $24$ questions per target set; the unrestricted
rerun reproduces each of those settings' cells exactly. On the construction-disjoint subset every
arm loses $2$--$4$ points of absolute accuracy \emph{including the random-pick
reference} ($35.1\!\to\!32.0$, $66.5\!\to\!63.4$, $70.9\!\to\!67.0$,
$78.8\!\to\!75.5$, $60.9\!\to\!57.1$ across the five settings), i.e.\ the removed
questions are easier, not leaked. The quantity the table actually reports --- the
gain over a random pick --- does not shrink: averaged over the five settings it is
$+5.2$ for \jsf{} on all $120$ questions and $+6.4$ on the disjoint $96$, and
$+4.0$ versus $+3.9$ for R64. Per setting, \jsf{}'s advantage is preserved or larger
in four of five ($+6.9\!\to\!+9.7$ at Low, $+1.6\!\to\!+1.9$ at Medium,
$+6.5\!\to\!+9.2$ on Qwen-Thinking, $+4.7\!\to\!+6.1$ on Qwen-Instruct) and smaller
at High ($+6.6\!\to\!+5.2$); the one arm that turns negative is R64 at Low effort
($+0.8\!\to\!-1.4$), which is the setting where Table~\ref{tab:bo64transfer}
already shows it barely clearing the reference.

\subsection{RQ4: Prefix Score, Controller and Baselines}
\label{app:rq4proto}

\paragraph{Prefix score.}
Windows are $256$ tokens with stride $256$. The target is $1-\text{outcome}$, i.e.\
the run will fail to deliver a correct answer (wrong answer or no answer), and the
predictor is a histogram gradient-boosted classifier ($150$ iterations, learning
rate $0.1$, at most $15$ leaves, minimum $200$ samples per leaf, $\ell_2$
regularization $1.0$, seed $0$), trained separately for each model and effort
setting over four question-level folds. The \jsf{} arm scores the window-mean
readout ($64$ features). The R64 arm first fits a least-squares map (ridge $10^2$)
from that window's model-specific routing spectrum ($99$ dimensions for gpt-oss-20b,
$387$ for Qwen3-30B-A3B) onto the window-mean readout
\emph{inside the training fold}, then scores the reconstruction --- so nothing but
routing is read at deployment.

\paragraph{Controller and replay.}
Cut when $S_t=\max(0,S_{t-1}+s_t-\kappa)$ exceeds $\tau$ and at least $512$ tokens
have been generated, with $\kappa$ the median training-fold window score and
$S_0=0$; $\tau\in\{0.25,0.5,1,2,4\}$ and the restart budget
$K\in\{2,3,4,6\}$. After $K$ cuts the next sample is allowed to run to completion.
Generation caps are $32{,}768$ tokens for gpt-oss and $32{,}000$ for Qwen.
Replay draws attempts from a per-question random
permutation of that question's pool without replacement --- $50$ permutations per
training question, $200$ per test question --- and if an attempt's natural length is
within its cut time it completes and is scored, otherwise the cut time is charged
and the next attempt begins. One unit of cost is the mean cost of the first run of
each permutation, i.e.\ letting one rollout finish. This paper uses two reporting protocols. The \emph{nested} protocol of
Table~\ref{tab:cusum} runs four outer question folds; within each fold it evaluates
every $(\tau,K)$ on the training questions only, brackets the target cost $c$ with
the two adjacent points $A,B$ of the training Pareto front and forms the randomized
mixture that plays $B$ with probability $(c-c_A)/(c_B-c_A)$, so the expected training
cost equals $c$; that fixed mixture is then executed on the held-out questions with no
maximum taken, and both the test accuracy and the realized test cost are reported.
Because nothing is selected on the questions being scored, the absolute levels are
not optimistic, and the realized costs stay close to their target and close to each
other (across both models, $1.42$--$1.92\times$ for \jsf{} against
$1.42$--$1.79\times$ for permSib at targets $1.4$--$2\times$); the exception is permAll, whose configurations saturate at
$1.18$--$1.31\times$, which is why we attribute against permSib and not against it. If
no training configuration is that cheap the fold falls back to an uninterrupted
rollout. Intervals throughout are $2{,}000$-resample bootstraps over
questions of the paired difference; they cluster questions but not permutations.

\paragraph{Baseline arms.}
Three zero-information controls pass through the identical CUSUM machinery, and
\emph{permSib} is the one every attribution uses, because it alone matches the
score's distribution and dynamics: it permutes whole score sequences between sibling
rollouts of the same question ($sd=0.221$, $ac_1=0.42$, window AUC $0.679$ against the
readout's $0.720$), preserving marginal, autocorrelation and question-level
information and destroying only branch-level information. The i.i.d.\ score is
$\mathrm{U}[0,1)$ per window (seed $5$), with within-run $sd=0.288$ and lag-one
autocorrelation $-0.01$ against $0.202$ and $0.37$ for the readout; \emph{permAll}
permutes globally (window AUC $0.48$) and its selected configurations saturate below
the cost targets. The confidence baseline follows DeepConf
\citep{fu2025deepconf}, which filters reasoning traces on model-internal confidence
and needs no extra training. We implement its windowed confidence signal, the negative mean of the top-$20$
token log-probabilities the sampler already stores over the trailing $256$-token
window, and none of the method's other components. In Table~\ref{tab:cusum} that scalar is read by the \emph{same} trained head and
the same CUSUM as every other arm, which is what makes the comparison a comparison
of signals. It needs token log-probabilities but no
internal state.

\subsection{Glossary}
\label{app:glossary}

\paragraph{Terms used above.}
\emph{Vocabulary direction}: a row of $D$, i.e.\ one token's direction in the
lens-projected space. \emph{Semantic frame}: the matrix $A=[a_1\,\cdots\,a_{64}]\in\mathbb{R}^{d\times64}$ whose columns are the axes.
\emph{State marker}: an axis whose neighbours are digits, script fragments or
register cues rather than content words; read as a position on the manifold rather
than literally. \emph{Echo hazard}: for digit axes, the risk that a reading merely
reflects digits present in the emitted text, which is why digit axes are never
named for content. \emph{Write-vector advantage}: an expert's projection onto a target
direction minus the mean projection of the experts it displaces.

\end{document}